\RequirePackage{fix-cm}
\documentclass[smallextended]{svjour3}       
\smartqed  
\usepackage{graphicx}
\usepackage{subfig}
\usepackage{url}
\usepackage{numprint}
\begin{document}

\title{Snapture - A Novel Neural Architecture for Combined Static and Dynamic Hand Gesture Recognition 
}


\author{Hassan Ali*        
        \and
        Doreen Jirak
        \and
        Stefan Wermter
}

\authorrunning{Ali et al.} 

\institute{
	H. Ali, S. Wermter \at
		University of Hamburg \\
		Knowledge Technology \\
		Department of Informatics \\
        Vogt-K\"{o}lln-Str. 30 \\
        22527 Hamburg \\
        Germany\\
        \emph{*Corresponding author:}\\
		\email{7ali@informatik.uni-hamburg.de} 
	\and
	D. Jirak \at
        IIT Central Research Labs Genova \\
		Robotics Brain and Cognitive Sciences \\
		Via Enrico Melen 83 \\
		16152 Genova \\
		Italy
}

\date{Received: date / Accepted: date}

\maketitle

\begin{abstract}
As robots are expected to get more involved in people's everyday lives, frameworks that enable intuitive user interfaces are in demand. Hand gesture recognition systems provide a natural way of communication and, thus, are an integral part of seamless Human-Robot Interaction (HRI). Recent years have witnessed an immense evolution of computational models powered by deep learning. However, state-of-the-art models fall short in expanding across different gesture domains, such as emblems and co-speech. 
In this paper, we propose a novel hybrid hand gesture recognition system. Our \emph{Snapture} architecture enables learning both static and dynamic gestures: by capturing a so-called \emph{snapshot} of the gesture performance at its peak, we integrate the hand pose along with the dynamic movement. Moreover, we present a method for analyzing the motion profile of a gesture to uncover its dynamic characteristics and which allows regulating a
static channel based on the amount of motion. Our evaluation demonstrates the superiority of our approach on two gesture benchmarks compared to a CNNLSTM baseline. We also provide an analysis on a gesture class basis that unveils the potential of our \emph{Snapture} architecture for performance improvements. 
Thanks to its modular implementation, our framework allows the integration of other multimodal data like facial expressions and head tracking, which are important cues in HRI scenarios, into one architecture. Thus, our work contributes both to gesture recognition research and machine learning applications for non-verbal communication with robots.  

\end{abstract}

\clearpage

\section{Introduction}
\label{sec:intro}


Gestures are a form of non-verbal communication prominently used in day-to-day human communication. Additionally, they have become a fundamental part of human-robot interaction (HRI). It is common in the literature to categorize gestures as static and dynamic. Static gestures portray particular meanings through hand postures. They can substitute words or be used in harmony with them in the form of signs or emblems. These gestures 
can be recognized through a precise interpretation of the emphasized hand shape and spelled out finger arrangements~\cite{siddharth2015}. In contrast, a dynamic 
gesture has a temporal aspect articulated through the movement of the hand. Therefore, recognizing it requires the employment of a different set of techniques, e.g., segmenting and tracking the moving body limb~\cite{anwar2019}.

However, such distinction between gesture types might overlook some of their unique characteristics. More specifically, hand pose is essential for recognizing gestures that share a similar motion path. For example, the gesture commands ``stop'' and ``go forward'' have an identical motion with the arm moving from the body side and extending to the front. However, the specific meaning of each command can be distinguished by observing their unique hand and fingers arrangement (open palm vs. extended finger). Furthermore, a precise interpretation of the unique characteristics of each hand gesture is desired for a smooth HRI experience. This becomes more vital in critical robot applications, e.g., the medical or industrial domains in which the confusion between gestures might have severe consequences, such as safety risks, in case of misinterpretation of a robot command.

This kind of precise interpretation is challenging for approaches that use RGB data only. Recently, we have witnessed a 
rise of multimodal data, such as depth and audio, provided through intelligent sensors. Various multimodal challenges and datasets were proposed recently~\cite{escalera2018}. However, these devices impose certain operational conditions~\cite{wang2018} limiting their flexibility. Therefore, RGB-based approaches are still desirable~\cite{chakraborty2017}. In addition to their convenience, they are potentially compatible with low-resource systems, such as robots. Furthermore, they facilitate the reproduction of results, especially as reproducibility issues related to deep learning are getting more attention by the scientific community~\cite{renard2020}. Although recent developments were triggered by the deep learning trend with networks like 3DCNN, ResNet, and Inception V3, dynamic gesture recognition is still a challenging task.

State-of-the-art vision-based approaches are challenged by factors such as \emph{indistinctive} and \emph{subtle} movements~\cite{amsterdam2021}. \emph{Subtle} movements refer to the small motion of the hand and fingers at the peak with no arm movement. On the other hand, \emph{indistinctive} movements mean that multiple gestures follow a very similar path of motion. One limitation prominent in various state-of-the-art approaches is that they rely on the motion path~\cite{asadi2017}. Consequently, some approaches lack in the consideration of hand details, i.e., the exact hand shape and finger arrangements. This leads to misclassifications between gestures with similar motion properties. Moreover, it 
is worth inspecting whether integrating hand details into the classification would refine the performance of such models.


In this study, we propose a modular RGB-based approach called \emph{Snapture}. It aims to address the issues of \emph{indistinctive} and \emph{subtle} movements in dynamic gesture recognition systems. Our architecture is an extension of the CNNLSTM~\cite{tsironi16} network and is evaluated in the domains of robot commands and co-speech gestures. This study is organized as follows: we present our literature review of some recent gesture recognition frameworks. Then, we describe the used datasets and our proposed method for analyzing gesture motion profiles. Next, we discuss the CNNLSTM architecture and our proposed hybrid gesture recognition framework called \emph{Snapture}. Then, we compare the performance of both models. Finally, we conclude with a discussion and highlight some potential directions for future research.


\section{Related Work}
\label{sec:related_work}
It is a common a step in vision-based hand gesture recognition systems to perform hand segmentation, i.e., extracting the moving hand from the background, before feeding the data into the learning model. In this context, Tsironi et al.~\cite{tsironi16} propose a pre-processing technique called the \emph{differential image} algorithm in which grayscale frames are consecutively subtracted. The processed data passes through a \emph{CNNLSTM} architecture, responsible for the implicit feature extractions and motion tracking of the hand across the time step sequences. The paper reports an accuracy of $\sim$0.92\ of the approach over the \emph{GRIT} dataset~\cite{tsironi17}. Despite having promising results, the study falls short in the correct classifications of gestures that share similar movement patterns. The authors report confusion between the ``hello'' and ``no'' classes, which both follow almost an exact motion path, i.e., \emph{indistinctive} movements. Most subjects perform ``hello'' and ``no''  using an open palm and an extended index finger, respectively. Therefore, the distinction lies in the specifics of the handshape and finger arrangements, which is not considered by the author's proposed approach. However, this does not highly influence the overall model's performance since it is limited to a small subset of gesture classes. In addition to this observation, the dataset used for evaluation is small (a total of 543 sequences over ten classes). Thus, it may not give a clear picture of the actual capabilities and limitations of the architecture. On the other hand, it is beneficial to assess the robustness of this approach against another gesture domain in which the correlation in motion across gestures is more pronounced. In our study, we evaluate this approach in the context of co-speech gestures. 

Moreover, it was reported in the literature that the performance of models could noticeably change based on the gesture vocabulary used for evaluation. Auge et al.~\cite{auge2020} demonstrate that the accuracy of their Numenta's Hierarchical Temporal Memory (HTM) approach drops drastically from 0.95 to 0.7 when evaluated on a subset of gestures. This is done by trimming their dataset of radar data from ten to five classes. Considering this observation, the problem is simplified by ruling out gestures with \emph{subtle} finger movements. This gives another clue on the non-triviality of distinguishing such particular hand and finger movements and their influence on the overall performance of models.

Similar to the CNNLSTM mentioned above, the loss of hand details is also reported in the work of Canuto et al.~\cite{canuto20}. In their RGB-based approach, the authors propose a motion representation technique called \emph{RGB star}. This method encodes color information, allowing compatibility with a broad spectrum of modern pre-trained CNN frameworks using transfer learning. 
In the presented approach, each gesture sequence is divided into three equal parts corresponding to the \emph{pre-stroke}, \emph{stroke}, and \emph{post-stroke} gesture's stages as defined by Kendon~\cite{kendon2011}. The algorithm generates a motion representation for each said part, further merged using the frame's R, G, and B channels. The pre-processed data is fed into a feature extraction model using pre-trained ResNet50 and ResNet101 CNNs. The features are further weighted using a soft-attention mechanism, while a final classification is accomplished using a two-layered feed-forward network. Similar to our work, the approach is evaluated in both the robot command and co-speech gesture domains. An accuracy of $\sim$0.98 and $\sim$0.95 is reported by the paper on the GRIT~\cite{tsironi16} and Montalbano~\cite{escalera2014} datasets, respectively. Despite the astonishing performance, the loss of hand details leads to 
confusion between multiple Italian gestures: ``noncenepiu'', ``ok'', ``freganiente'', and ``prendere''. These movements share a similar path with the hand raised over the elbow. Additionally, some of them, e.g., ``noncenepiu'', are \emph{subtle}. Furthermore, the architecture is overly complex, especially considering that the GRIT is a small-scale dataset, as discussed earlier. 
However, the results of this paper further confirm that the issues of \emph{indistinctive} and \emph{subtle} gestures are not trivial. The authors hypothesize that they can be addressed by integrating the hand information into the system. However, it remains an open question whether that would improve the performance.

The stated hypothesis is supported by the work of Wu et al.~\cite{wu2016}. By fusing RGB and depth data alongside a skeleton modality, the authors demonstrate an increase in performance concerning gestures with similar motions. The authors propose an approach called Deep Dynamic Neural Network (DDNN), which consists of three neural networks, corresponding to the following modalities: RGB, depth, and skeleton. RGB and depth data is fed into a 3DCNN, while skeleton data passes through a Deep Belief Network (DBN). The learned features of these networks are fused and further fed into a feed-forward network that produces an emission probability at each time step. A Hidden Markov Model (HMM) is responsible for the temporal modeling gestures and pauses using a set of defined states. Each observation is classified by calculating the most probable path using the Viterbi algorithm. A score of 0.816 is reported on the Montalbano dataset~\cite{escalera2014} with the Jaccard index. The results hint at the importance of integrating RGB data when preserving the hand pose of gestures. The authors report less confusion regarding \emph{implicit} Montalbano class movements when considering RGB and depth modalities. However, the approach does not consider Kendon's model~\cite{kendon2011} of co-speech gestures, which states that the gesture's hand pose is uncovered during the \emph{stroke} phase. 
Hence, the hand pose is treated equally at all gesture phases. Consequently, some gestures are considered similar based on resemblance at the beginning or end of the motion, i.e., during the \emph{rest phase}. Furthermore, their model is computationally intensive and requires long training times of five days. Thus, its robotic applications might be limited.


One approach that attempts to integrate static and dynamic recognition is the architecture of Mazhar et al.~\cite{mazhar2021}. In their CNNLSTM-based framework called StaDNet, they propose an architecture consisting of two Inception V3 CNNs. Each CNN is responsible for the spatial extraction of the left and right hands. The authors claim that by cropping the CNN input to the hand and removing the background, the framework can learn \emph{subtle} movements. The temporal learning is carried out using an LSTM network in which the features from multiple modalities are fused. Besides RGB, a 2D body skeleton model is extracted using OpenPose\footnote{\url{https://github.com/CMU-Perceptual-Computing-Lab/openpose}}. Additionally, the approach includes depth estimators extracted using a Kinect sensor. These estimators highlight the area of interest, i.e., the hand. This model requires two datasets for training: one static and one dynamic. The accuracy of 0.8675 and 0.989 is reported on the Chalearn 2016 and OpenSign datasets, respectively. However, the requirement of two independent sets of gesture vocabulary implies that the architecture learns them separately. Thus, the model can not seamlessly classify a gesture based on its dynamic and static characteristics in contrast to the authors' claim. Similarly, despite the model's ability to classify new samples using RGB data only, it requires multiple modalities for training. This contradicts the author's claim of a pure RGB-based framework.

\section{Datasets and Motion Profiles}
\label{sec:datasets}
As mentioned in our literature review, the performance of gesture recognition frameworks might be influenced by the choice of the gesture domain. For example, co-speech gestures are more natural than robot commands. Therefore, they might incorporate higher chances of encountering \emph{indistinctive} and \emph{subtle} movements. Therefore, our choice of dataset spans across multiple gesture domains, i.e, the contexts of robot commands and co-speech gestures. In this section, we present the two datasets used to evaluate our framework. Due to the substantial difference between the domains of the datasets, we find it crucial to carry out a temporal analysis of the gesture sequences. Analysing gestures would provide insights into how the motion changes over their time span. Late in this section, we will describe our SSIM-based method for analyzing the motion profile of gestures. We present a motion analysis of several gestures from the said datasets and highlight some of the key distinctions. Based on that, we define two kind of gestures, \emph{paused} and \emph{repeating pattern} gestures, which we also define.

\subsection{The GRIT Robot Commands Dataset}
\label{sec:grit_dataset}
In the context of robot commands, the ``Gesture commands for Robot InTeraction" (GRIT)\footnote{\url{https://www.inf.uni-hamburg.de/en/inst/ab/wtm/research/corpora.html} 
}~\cite{tsironi17} is one of the few publicly available dynamic gesture datasets. The corpus contains 543 \emph{isolated gestures} distributed over nine gesture classes and recorded with the help of six participants. 
However, most movements are designed to have a unique path of motion. An exception is present in the case of classes ``hello'' and ``no'' which are \emph{indistinctive movements}, as discussed previously. 
The recorded gestures vary in length, which is evident for gestures 
such as ``circle'' and ``turn''. Similarly, the subjects were not given any instructions on how to perform the gestures nor which hand to use. Consequently, it becomes more challenging to capture the hand pose of some gestures, especially with the camera's relatively low frame rate and resolution, as we will discuss later. The dataset was collected under lab-controlled settings with a plain white background. All subjects have similar lighting conditions with no noise in the surrounding.

\subsection{The Montalbano V1 Co-Speech Dataset}
\label{sec:montalbano_dataset}
The Montalbano dataset is a publicly available corpus in the domain of co-speech gestures. This dataset was collected as part of the \emph{ChaLearn\footnote{\url{http://chalearnlap.cvc.uab.es/} 
} Looking at People challenge}~\cite{escalera2014}. It contains around \numprint{14000} Italian gestures spreading over 20 gesture classes. We use this dataset to evaluate our approach in the context of gestures ``in the wild''. Although each recorded video contains one subject only, the data is collected in various day-to-day human environments, such as offices and lecture halls, with different noisy backgrounds. The Montalbano dataset tackles the task of multimodal gesture spotting. Therefore, it includes multiple sensory data, e.g., depth, user index, skeleton, and RGB images. However, we only use RGB data due to the advantages of reproducibility and portability that vision-based approaches provide. Since the gestures were recorded continuously with little to no pause between them, we convert them into \emph{isolated gestures} by identifying the start and end of each movement sequence. The variation in length is also pronounced in this dataset, which is partly due to the high number of subjects ($\sim$50). 
We make the annotations created for isolating the gestures and source code of the experiments presented in this work publicly available\footnote{\url{https://github.com/hassanali-90/snapture/}}.

\subsection{Motion Profile Analysis}
\label{sec:motion_profile_analysis}
Our approach for tackling the issues concerning \emph{indistinctive} and \emph{subtle} movements relies on fusing the hand motion and pose. Therefore, we refer to our architecture (described in the next section) as \emph{hybrid}. Motion features can be extracted by exploiting the temporal information across the consecutive frames. The hand pose is most interesting at the \emph{stroke} phase of co-speech gestures, as described by Kendon~\cite{kendon2011}. However, the relationship between frames and \emph{stroke} in the targeted datasets is not clear. Therefore, we analyze hand gesture sequences in the studied datasets to uncover the characteristics of their dynamics in terms of motion and pause. Due to the lack of approaches for gesture analysis, we utilize the structural similarity index measure (SSIM)~\cite{wang2004} as a metric for the similarity between consecutive frames. We carry out the calculation as follows:

\begin{equation} 
\label{eq:ssim_1}
SSIM(x, y) = \frac{(2\mu_{x}\mu_{y} + C_{1})(2\sigma_{xy} + C_{2})}{(\mu_{x}^2 + \mu_{y}^2 + C_{1})(\sigma_{x}^2 + \sigma_{y}^2 + C_{2})}\,,
\end{equation}

\begin{equation} 
\label{eq:ssim_2}
C_{1} = (K_{1}L)^2\,,
\end{equation}

\begin{equation} 
\label{eq:ssim_3}
C_{2} = (K_{2}L)^2\,,
\end{equation}

where $x$ and $y$ are spatially local windows of the input frames. $\mu_{x}$ and $\mu_{y}$ represent the mean intensity of $x$ and $y$, respectively. Similarly, $\sigma_{x}$ and $\sigma_{y}$ denote the standard deviation. Stability constants $C_{1}$ and $C_{2}$ help avoid a division by zero and are calculated using the pixel range $L$ (255 for 8-bit grayscale frames) and positive values much smaller than 1, i.e., $K_{1}$ and $K_{2}$. 

Using the first frame as a reference, we can quantify the amount of motion and pause across the gesture time span. By inverting the equation, we can express 
change across frames. We refer to that as the \emph{Inverted SSIM (ISSIM)}: 

\begin{equation} 
\label{eq:inverted_ssim}
ISSIM = 1 - SSIM (I_{i}, I_{0})\,,
\end{equation}

where $I_{i}$ and $I_{0}$ denote the grayscale frames at time steps $i$ and $0$, respectively.



\begin{figure}
  \includegraphics[width=1\textwidth]{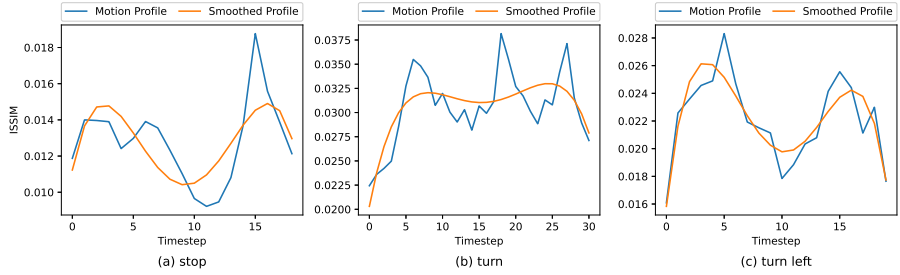}
\caption{The motion profile of GRIT gestures ``stop'', ``turn left'' and ``turn''. ``stop'', ``turn left'' are \emph{paused} at their peak, while ``turn'' is with a \emph{repeating pattern} due to the continuous intensity across its time span.}
\label{fig:grit_motion_profile} 
\end{figure}

Due to their design as robot commands, we observe two variations of movements in the GRIT dataset based on the analyzed motion profile. \emph{Paused} gestures include a pronounced period of pause around the gesture peak. For example, ``stop'' (cf. Fig.~\ref{fig:grit_motion_profile} (a)) and ``turn left'' (cf. Fig.~\ref{fig:grit_motion_profile} (c)) lack motion around the peak since participants hold their hand briefly in a fixed position. In contrast, in gestures, such as ``turn'' (cf. Fig.~\ref{fig:grit_motion_profile} (b)), subjects continuously repeat a circular pattern across the gesture's time span. We refer to these movements as \emph{repeating pattern} gestures. As we will see later, these unique characteristics of motion and pause of each gesture influence the design of our approach. 

\begin{figure}
  \includegraphics[width=1\textwidth]{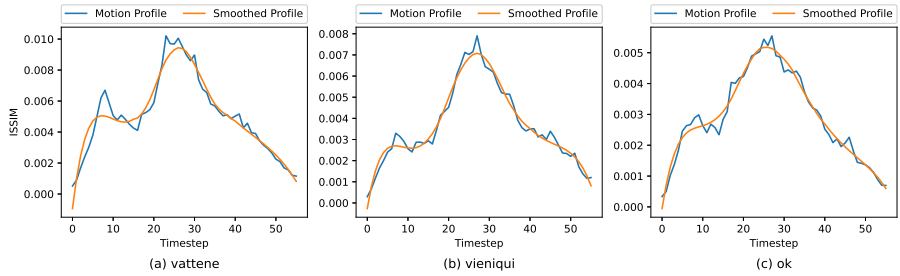}
\caption{The motion profile of Montalbano gestures ``vattene'', ``vieniqui'' and ``ok''. These co-speech gestures follow Kendon's model~\cite{kendon2011}, hence, they start and end with low intensity and have a clear peak around the midpoint of the timeline.
}
\label{fig:montalbano_motion_profile} 
\end{figure}

In contrast to the GRIT dataset, the Montalbano gestures are co-speech. Thus, they follow Kendon's~\cite{kendon2011} relational model of gesticulation and concurrent speech. The intensity of the movement starts and ends gradually with a clear peak in between. In Fig.~\ref{fig:montalbano_motion_profile}, we see examples of the motion profile of Montalbano gestures.

\section{Snapture - Hybrid Gesture Recognition}
\label{sec:snapture}
\begin{figure}
    \centering
    \includegraphics[width=0.3\textwidth]{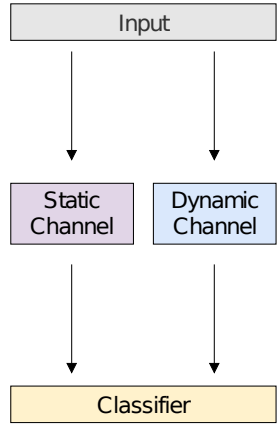}
\caption{An overview of the \emph{Snapture} framework. The architecture consists of a dynamic and static channels, fused into a final classifier. Thus, it performs a hybrid hand gesture recognition task.}
\label{fig:snapture_overview}
\end{figure}

One core concept of our hybrid recognition system is to combine the dynamic and static (movement and hand pose) aspect of gestures. The analysis of the GRIT and Montalbano gestures provides insights into their motion profiles. This information will be utilized when extracting the hand pose at the peak, as we will see later. In this section, we will describe our proposed approach called \emph{SNAPshot capTURE} (\emph{Snapture}). 
Our approach is an extension of the \emph{CNNLSTM}~\cite{tsironi16} architecture, and aims to find a solution to the problems of \emph{indistinctive} and \emph{subtle gestures}. This is done by integrating the hand details in a dynamic gesture recognition framework, thus performing a hybrid gesture recognition task. A simplified overview can be seen in Fig.~\ref{fig:snapture_overview}.


\subsection{Dynamic Channel}
\label{subsec:dynamic_channel}
\begin{figure}
    \includegraphics[width=1\textwidth]{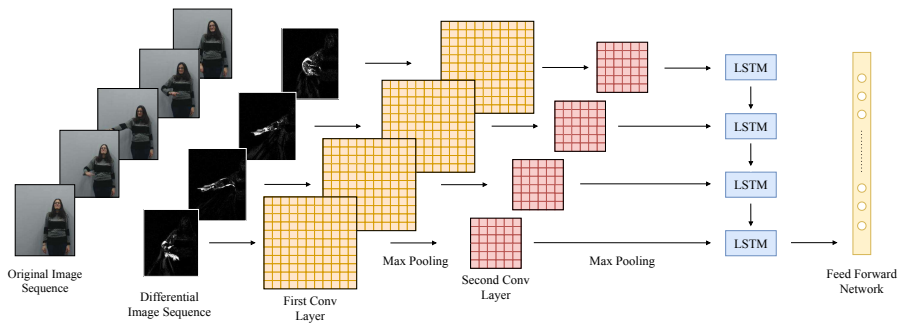}
\caption{The dynamic channel of \emph{Snapture} is a CNNLSTM network, consisting of two layers of CNN followed by a LSTM and feed forward network. The \emph{isolated gesture} input is pre-segmented using the \emph{differential image} algorithm. For clarity, we show only five frames and increase the contrast of the differential images.
}
\label{fig:snapture_dynamic_channel}
\end{figure}

The \emph{CNNLSTM}~\cite{tsironi16} architecture is an RGB approach, which has proven to work quite well classifying various motion patterns of robot commands. The CNNLSTM network consists of a 2-layer stacked convolutional neural network (CNN) followed by a long short-term memory (LSTM) network (cf. Fig.~\ref{fig:snapture_dynamic_channel}). The input frames represent segmented gestures, which means the moving hand is detected and extracted in the input by subtracting subsequent frames. This is calculated using the \emph{differential image} algorithm as described in \cite{tsironi16}:

\begin{equation} 
\label{eq:diff_images}
\Delta_{i} = (I_{i} - I_{i-1}) \wedge (I_{i+1} - I_{i})\,,
\end{equation}

where $\Delta_{i}$ and $\Delta_{i-1}$ are the segmented gesture input frames at the current and previous time steps, respectively. $I_{i-1}$, $I_{i}$ and $I_{i+1}$ denote the grayscale frames at time steps $i-1$, $i$ and $i+1$, respectively. $\wedge$ is the bitwise AND operator. Additionally, each input sequence represents an \emph{isolated gesture}, i.e., the sequence's start and end is known. 

As motivated earlier, we are interested in evaluating our \emph{Snapture} approach across multiple gesture domains, i.e., robot commands and co-speech gestures. The \emph{Snapture} architecture is an extension of the CNNLSTM model. We will use the CNNLSTM method as described in~\cite{tsironi16} as baseline for comparison. On the other hand, little is known so far about the performance of CNNLSTM in the context of co-speech gestures. 
Therefore, our evaluation provides some further insights into the performance of CNNLSTM using the Montalbano dataset. Our PyTorch\footnote{\url{https://pytorch.org/}} implementation of the CNNLSTM uses the same kernel size and number of filters of the CNN as the original proposal~\cite{tsironi16}. However, we tune the rest of the experimental settings. The stacked convolution layers have five and ten kernels of size 11x11 and 6x6, respectively. Each layer has a 1x1 stride, zero-padded input, and a hyperbolic tangent (Tanh) activation function. We chose the non-linearity using grid-search with the rectified linear unit (ReLU) as an additional candidate. A max-pooling layer of size 2x2 follows each convolution layer. Additionally, batch normalization is used after each convolution to reduce internal covariate shift~\cite{ioffe2015} and speed up the training. We initialize the CNN's weights with values from a uniform distribution~\cite{glorot2010}. The output of the last convolution layer is flattened and propagated through a feed-forward layer. 

Due to the input of \emph{isolated gestures}, each mini-batch has all the information needed for the network to produce a classification. Therefore, we opt to use a stateless LSTM. The LSTM's number of layers and neurons are selected using grid search. The optimal number of layers is 2 out of 1, 2, 4, and 8. The optimal number of neurons has resulted differently for the GRIT and Montalbano datasets. We chose 64 and 512 neurons for the GRIT and Montalbano datasets, respectively. We initialize the LSTM with weights from a uniform distribution with zero bias. After passing through dropout~\cite{pham2014}, 
the output of the LSTM is further propagated into feed-forward and softmax layers, producing a probability distribution over the gesture classes. The rest of the hyperparameters will be presented in a later section.

Due to the cell state of the LSTM, the CNNLSTM architecture's output can be configured in various ways. In the original CNNLSTM proposal, the authors have opted for a \emph{frame-level} configuration. Therefore, their model predicts a label for each time step. In contrast, we tune our model to produce a \emph{sequence-level} classification for each gesture sequence in its entirety. Therefore, our model requires no additional post-processing steps and fits the concept of capturing a \emph{snapshot} more intuitively. 

\subsection{Static Channel}
\label{subsec:static_channel}
This channel is responsible for capturing the specific handshape and finger arrangements through a so-called \emph{snapshot} at the gesture's peak. We detect and extract the gesture at the peak corresponding to the \emph{stroke} phase. This provides the hand pose information, which can be fused alongside CNNLSTM's motion learning outcome. As a result, our method integrates the characteristics of static and dynamic recognition systems. 

\subsubsection{Gesture Peak Detection}
\label{subsec:peak_detection}
\begin{figure}
    \subfloat[\centering Rest position]{{\includegraphics[width=0.18\textwidth]{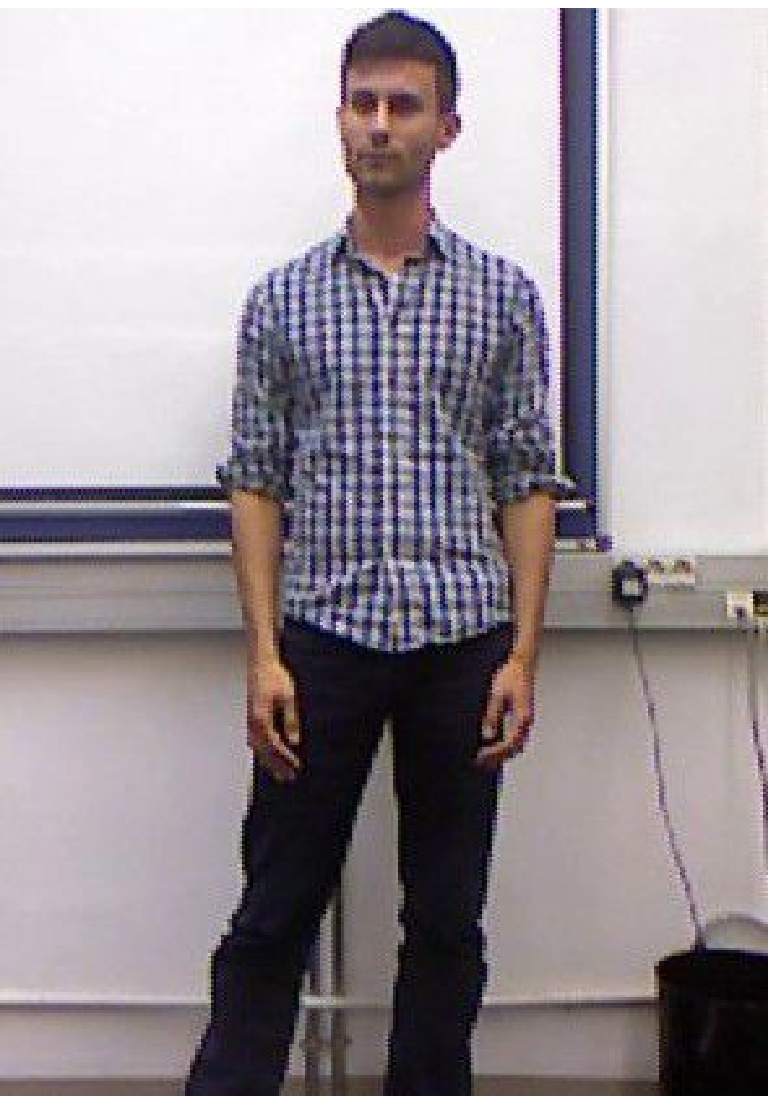} }}%
    \,
    \subfloat[\centering Pre-stroke]{{\includegraphics[width=0.18\textwidth]{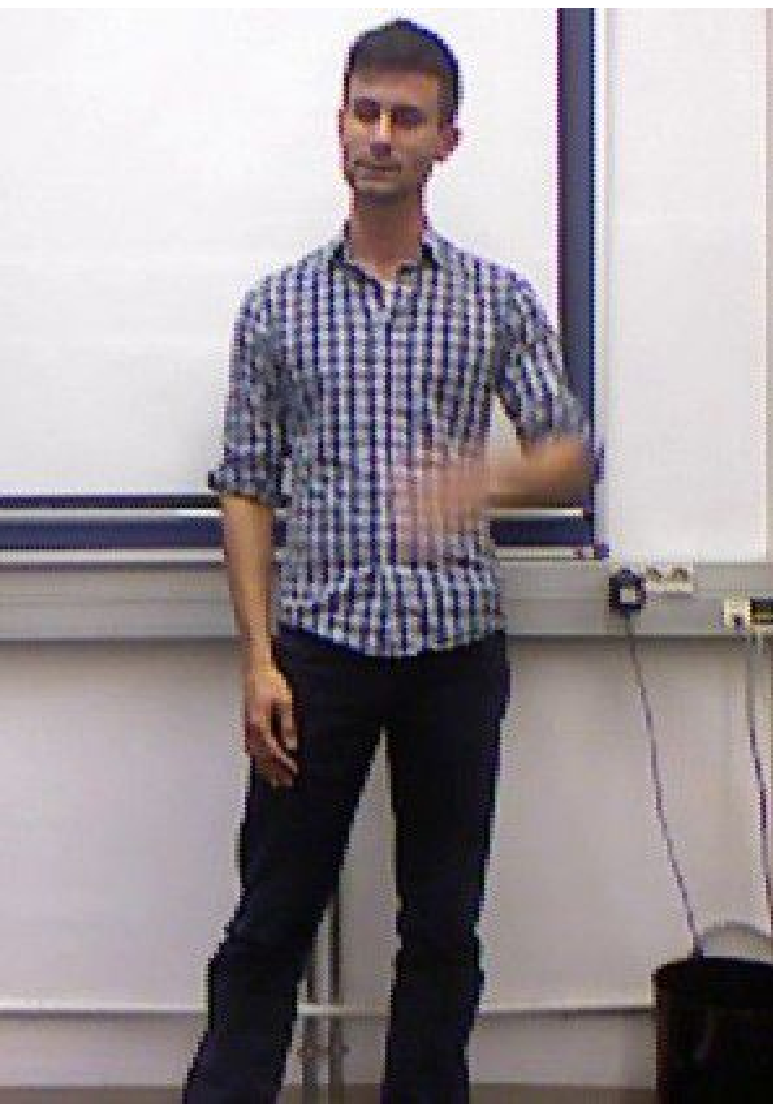} }}%
    \,
    \subfloat[\centering Stroke]{{\includegraphics[width=0.18\textwidth]{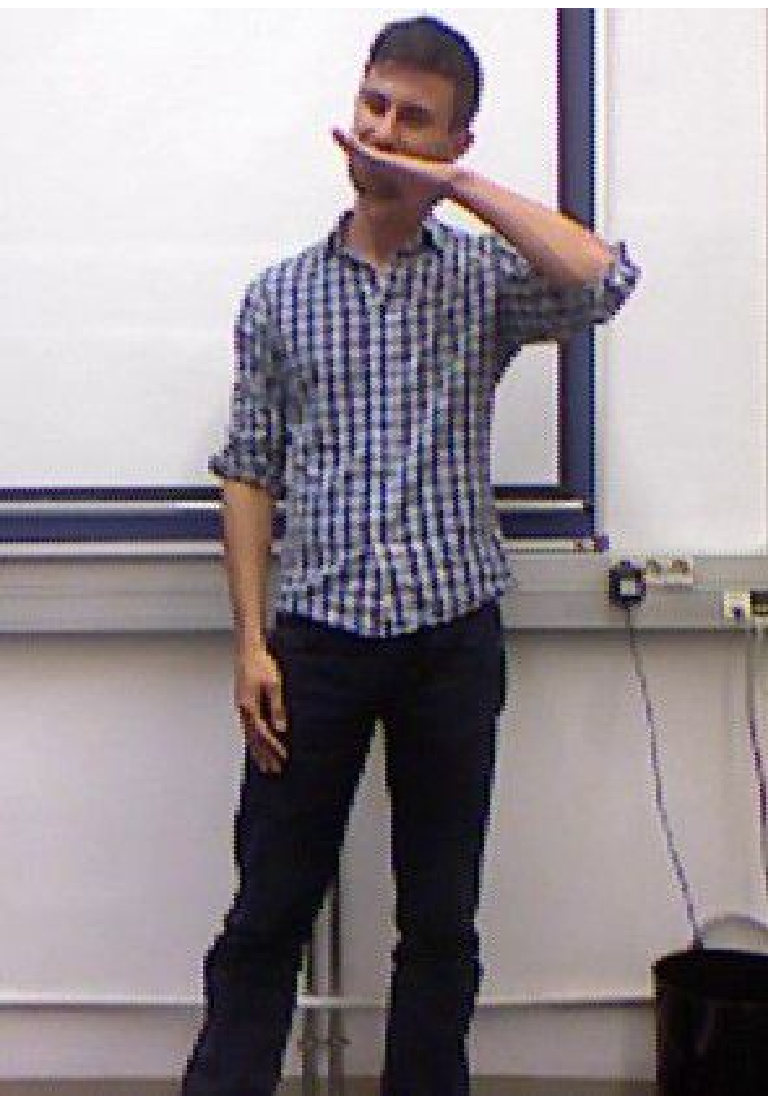} }}%
    \,
    \subfloat[\centering Post-stroke]{{\includegraphics[width=0.18\textwidth]{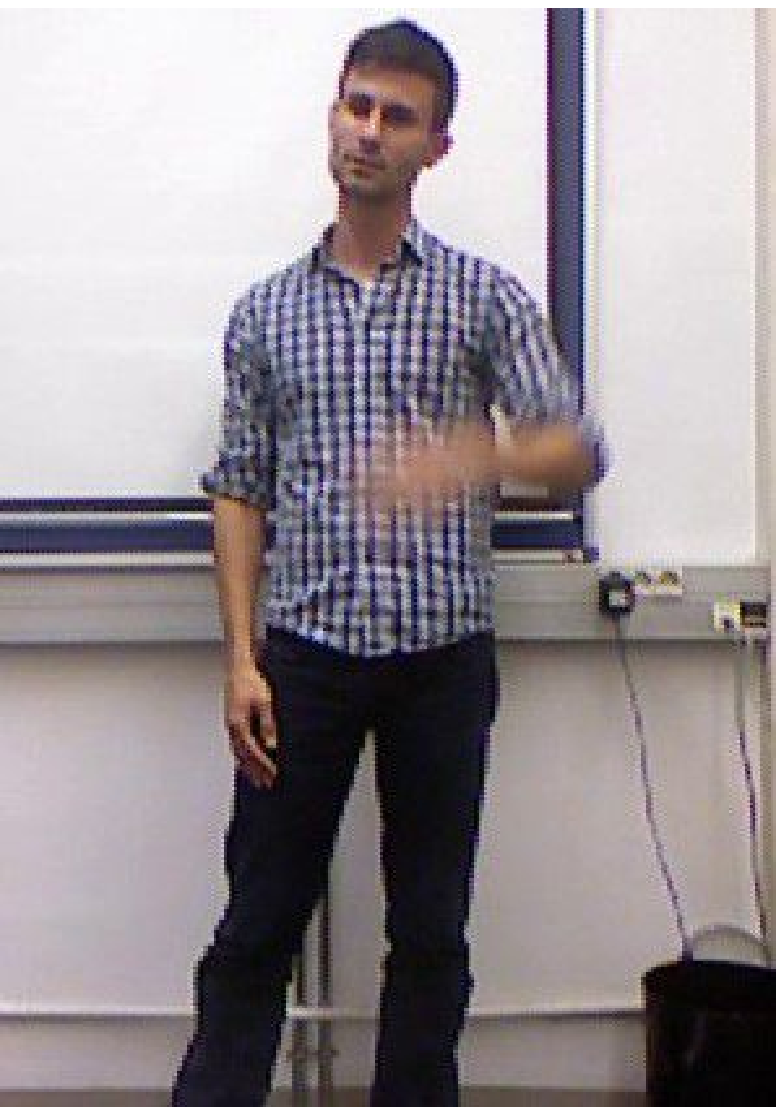} }}%
    \,
    \subfloat[\centering Rest position]{{\includegraphics[width=0.18\textwidth]{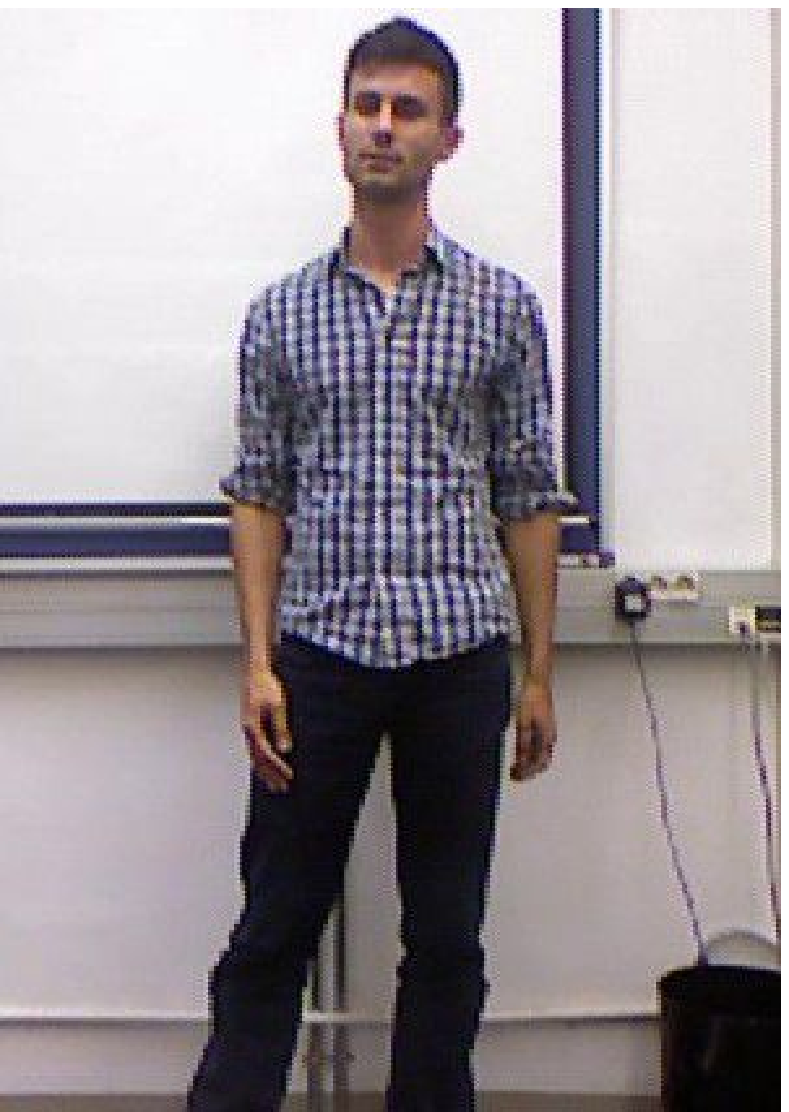} }}%
\caption{The five gesture phases of Kendon~\cite{kendon2011}. Each gesture starts with a \emph{rest phase}. In \emph{pre-stroke}, the limb moves from the rest position into the \emph{stroke} phase. The \emph{stroke} phase contains the most expressive information. 
In post-stroke, the limb moves away from \emph{stroke} back into \emph{rest phase}. 
}
\label{fig:kendon_phases}
\end{figure}

According to Kendon's~\cite{kendon2011} model of the relationship between gestures and concurrent speech, human gestures are described by five phases (cf. Fig~\ref{fig:kendon_phases}). Gestures start with a \emph{rest phase}, which represents a non-movement state of the arms. In the \emph{pre-stroke} or \emph{preparation phase}, a gradual intensity in motion of one or both arms starts to unveil. Next is the \emph{stroke} phase in which the gesture static characteristics, i.e., hand shape and finger configurations, completely unfold. These characteristics start to fade away in the \emph{post-stroke}, or \emph{retraction} phase as the intensity of motion gradually decreases. The gesture ends again with a rest phase. Montalbano gestures have a clear peak through the frames around the midpoint of the gesture sequence (cf. section~\ref{sec:montalbano_dataset}). Similar time steps are occupied by a pronounced pause in \emph{paused} gestures (cf. section~\ref{sec:grit_dataset}). Since the data consists of \emph{isolated gestures}, we identify the peak as the frame in the middle of the gesture sequence.

\subsubsection{Gesture Peak Extraction}
\label{subsec:peak_extraction}
We follow a skin detection technique to extract the hand from the rest of the frame. Our implementation uses Python 
and OpenCV\footnote{\url{https://opencv.org/opencv-4-5-3/}}. First, the face of the subject is detected and removed from the frame since skin detection techniques treat all visible skin of body parts in the frame equally. 
Next, the hand is segmented by converting into the orthogonal color space \emph{YCbCr}~\cite{hsu2002}: \emph{Y} representing the luminance, while \emph{Cb} and \emph{Cr} indicate the chromaticity. This is done to avoid the high correlation between luminance, hue, and saturation in RGB~\cite{qiu-yu2015}. Since various lighting conditions highly influence skin tones, we apply the threshold on chrominance only. We use the thresholds \emph{Cb}=[80, 120] and \emph{Cr}=[133, 173] proposed by Basilio et al.~\cite{basilio2011}. According to the authors, these threshold values are independent of skin tone. However, the datasets used in our study are limited in the diversity of skin tones. 
An additional step of background removal is applied to the Montalbano data using simple background subtraction. This is due to the complexity of recording scenes, unlike GRIT. Furthermore, we apply the connected component analysis, 
which describes the \emph{YCbCr} mask in terms of BLOBs. 
These objects are then sorted by size and position. Due to the noisy background in the Montalbano dataset, we filter out objects that do not belong to the foreground, calculated in the step of background removal.

\begin{figure}
    \centering
    \includegraphics[width=1\textwidth]{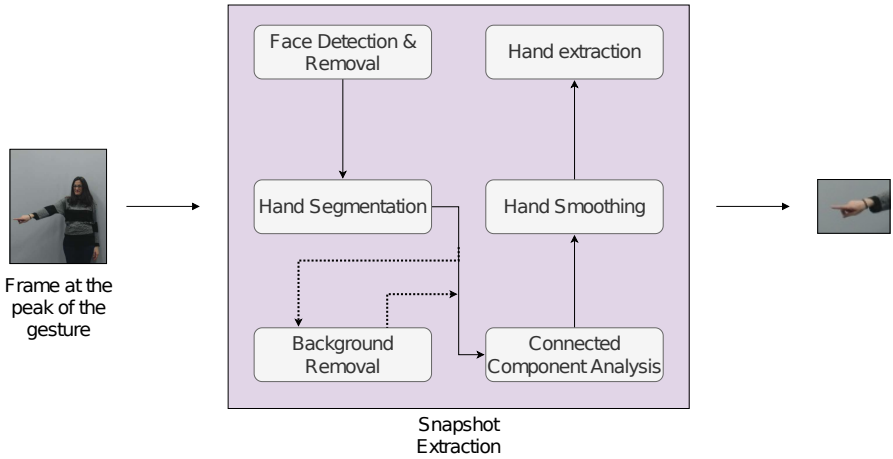}
\caption{The gesture peak extraction module of the \emph{Snapture} approach. Using a skin detection technique, the hand shape and finger configurations are extracted from a target frame at the gesture's peak. Background removal is only applied to the Montalbano gestures (dotted line).
}
\label{fig:snapshot_extraction}
\end{figure}

To avoid any subject's preferred hand assumption, we pick the higher object in the frame. As we observe in the data, all subjects have the hand performing the gesture in an upper position, while the other is usually in rest or slightly raised. For gestures requiring two hands, both always contain the same hand pose. Therefore, our algorithm has the flexibility of picking up either hand in this case. 
A step of hand smoothing is applied using erosion and dilation morphological transformations. However, we find that omitting this step does not influence the algorithm's output. Finally, an area around the detected hand is extracted from the original frame and further resized to 64x48 pixels matching the CNN input. The gesture peak extraction module is depicted in Fig.~\ref{fig:snapshot_extraction}.

\subsubsection{Static Channel Control}
\label{subsec:threshold_mechanism}

\begin{figure}%
    \centering
    \subfloat[\centering]{{\includegraphics[width=5cm]{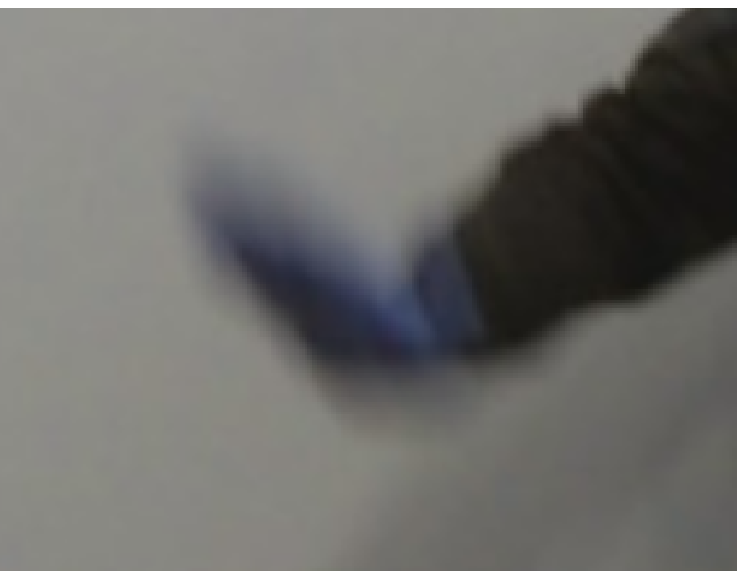} }}%
    \qquad
    \subfloat[\centering]{{\includegraphics[width=5cm]{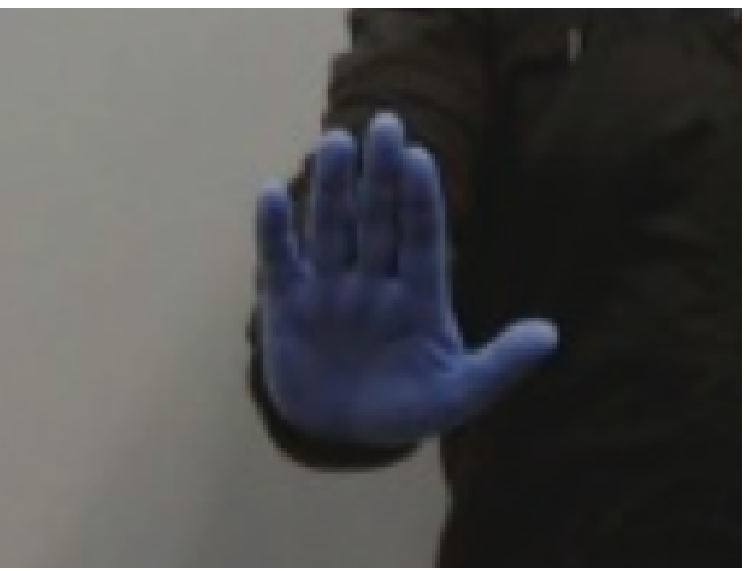} }}%
    \caption{ \emph{Repeating pattern} gestures, e.g., ``circle'' (a),  contain a \emph{blur} at the peak compared to \emph{paused} movements, e.g., ``stop'' (b). The blurry hand at the gesture's peak for highly dynamic movements is challenging for RGB-based approaches. We bypass this issue by regulating the static channel of our approach. }
    \label{fig:blur}%
\end{figure}

Despite the convenience of vision-acquired data, modest RGB cameras tend to have certain limitations. For example, they fail to capture the hand details when a rapid movement is present. This is caused by factors such as camera resolution and exposure time. Consequently, it leads
to a blurry hand in the frame (cf. Fig.~\ref{fig:blur}). This is pronounced for \emph{repeating pattern} robot commands due to the high dynamics of movement. This phenomenon raises a challenge to any vision-based approach due to the missing information and limited data source quality. However, we aim to address it by regulating the static channel based on the amount of motion contained in a gesture. More precisely, we integrate the extracted static information, i.e., the hand shape at the peak, only if the amount of motion lies below a threshold., i.e., the \emph{stroke} phase contains a pause sufficient for the \emph{snapshot} extraction.

\begin{figure}
    \centering
    \includegraphics[width=1\textwidth]{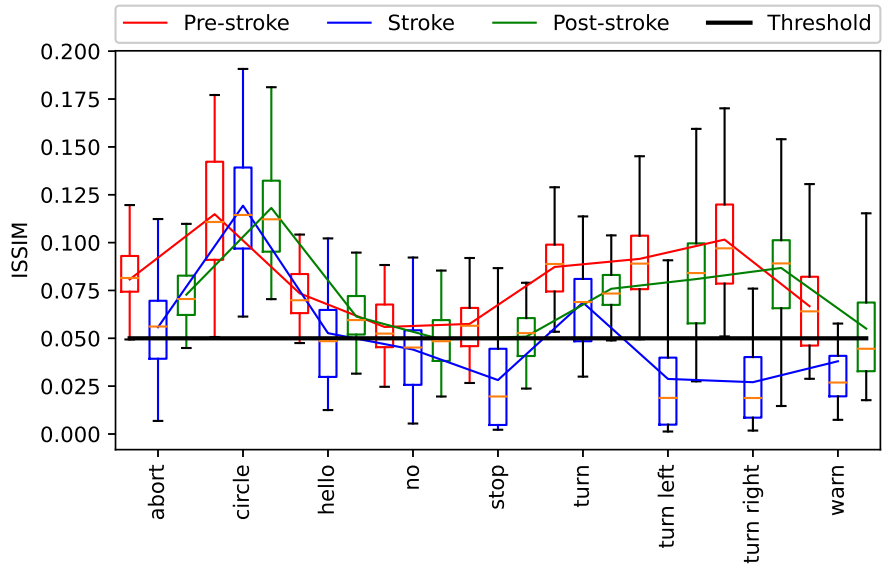}
\caption{Our motion analysis of the GRIT dataset after splitting into three parts: \emph{rest} to \emph{pre-stroke} phases, \emph{pre-stroke} to \emph{post-stroke} phases and \emph{post-stroke} to \emph{rest} phases. The second part contains more pause and facilitates capturing a \emph{snapshot}. The black line denotes our defined threshold for regulating the static channel.
}
\label{fig:grit_threshold}
\end{figure}

\begin{figure}
    \centering
     \includegraphics[width=1\textwidth]{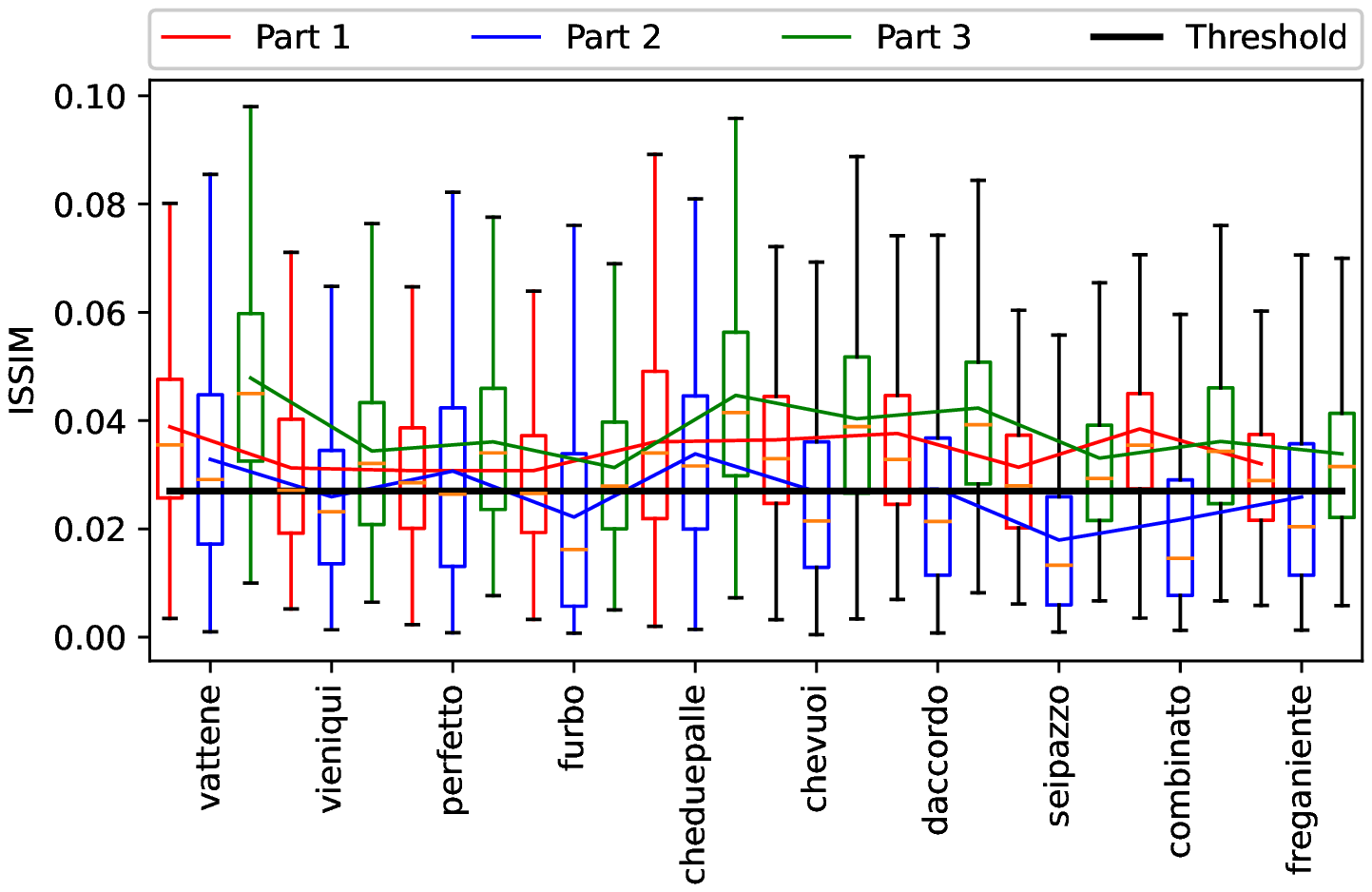}
    \includegraphics[width=1\textwidth]{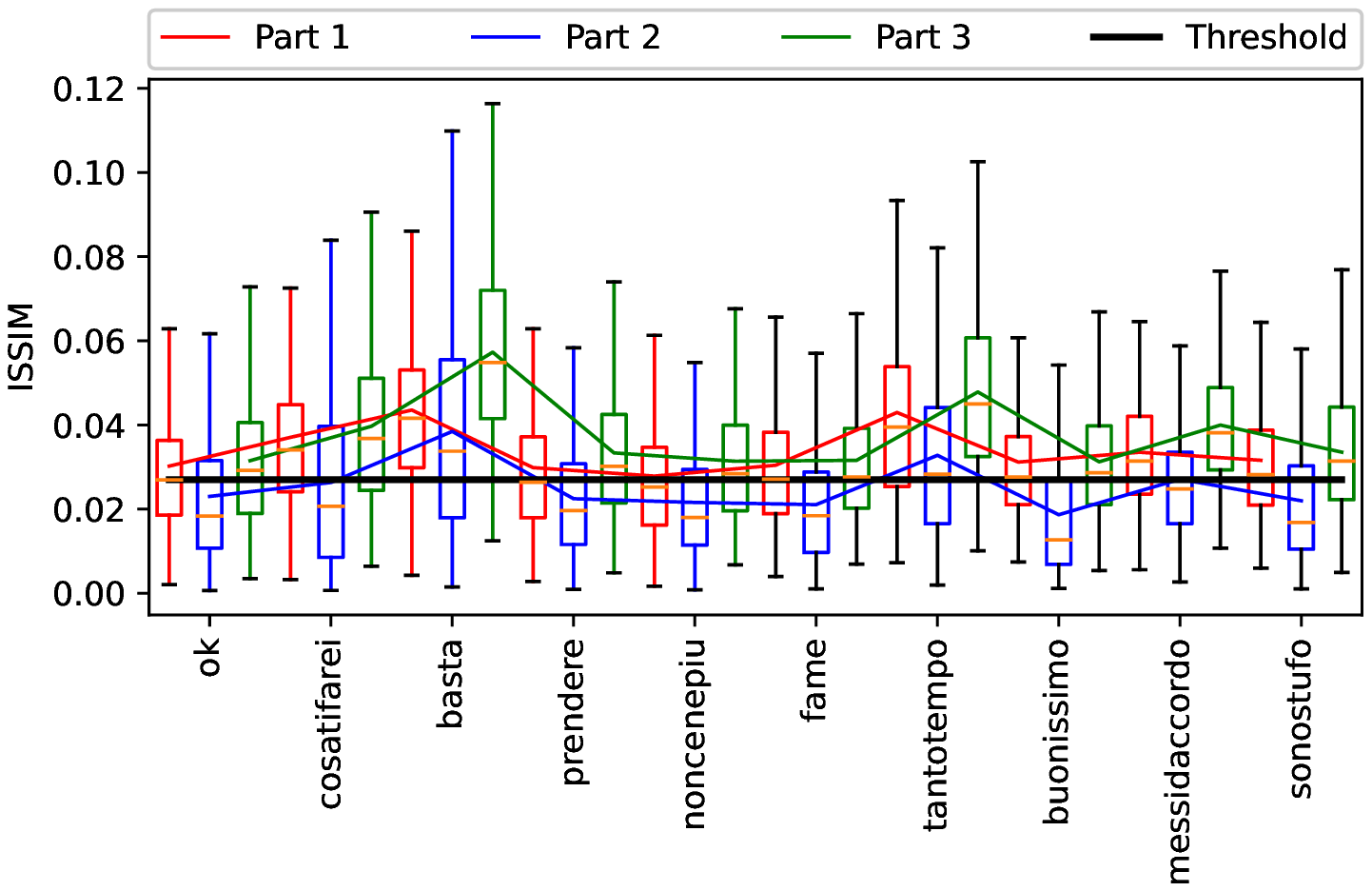}
\caption{Our motion analysis of the Montalbano dataset after splitting into three parts: \emph{rest} to \emph{pre-stroke} phases, \emph{pre-stroke} to \emph{post-stroke} phases and \emph{post-stroke} to \emph{rest} phases. Similar to GRIT, the second part contains more pause that facilitates capturing a \emph{snapshot}. Our defined threshold for regulating the static channel is denoted by the black line.
}
\label{fig:montalbano_threshold}
\end{figure}

We use the SSIM based approach presented in section~\ref{sec:motion_profile_analysis} as quantitative metrics for the amount of motion and pause. We split each \emph{isolated gesture} into three parts with equal number of frames: 1) the first part represents all the frames in the \emph{rest} and \emph{pre-stroke} phases, 2) the second part contains the frames in \emph{pre-stroke} and \emph{post-stroke} phases, 3) and the third part consists of all the frames consecutively from \emph{post-stroke} to \emph{rest} phases. We assume the three parts to be of equal length for simplicity. The three parts and our defined threshold are visualized for the GRIT (c.f. Fig.~\ref{fig:grit_threshold}) and Montalbano (c.f. Fig.~\ref{fig:montalbano_threshold}) datasets. The average amount of motion in part 2 is less than part 1 and part 3, which supports our choice of Kendon's~\cite{kendon2011} \emph{stroke} phase as the peak of the gesture. Furthermore, most samples of \emph{paused} gestures, such as: ``stop'', ``turn left'' and ``turn right'', lie well below the threshold due to their pronounced period of pause. In contrast, the intensity of motion is evidently high for \emph{repeating pattern} gestures, e.g., ``circle'' (cf. Fig.~\ref{fig:grit_threshold}). This goes well with our definition of \emph{repeating pattern} gestures (cf. section~\ref{sec:grit_dataset}) since the \emph{stroke} phase of these commands does not contain any distinct hand shape information. Thus, the shape of the hand plays a minimal role in the recognition, and by disabling the static channel for ``circle'' samples, we can reduce the influence of the drastic loss of sharpness in the input. On the other hand, the majority of Montalbano gesture classes contain pause facilitating the capturing of a \emph{snapshot}. The cut-off value lies around the median of each gesture class except for ``basta'' and ``cheduepalle'', which have an explicit arm movement to the side of the body, as we will discuss later.

\section{Experiments and Results}
\label{sec:experiments_results}
In this section, we present the experiments carried out in this study. In each experiment, we evaluate and compare the following: 1) a \emph{sequence-based} CNNLSTM model, which classifies gestures based on motion only and acts as a baseline for comparison, 2) our \emph{Snapture} architecture without the threshold mechanism, which predicts a class by integrating the handshape and motion, and 3) \emph{Snapture} with the threshold-controlled mechanism 
for regulating the static channel based on the sufficiency of pause to capture a \emph{snapshot}. We will refer to this model as \emph{Snapture\textsubscript{thold}}. We repeat the experiment in the contexts of robot commands and co-speech gestures using the GRIT and Montalbano datasets, respectively. The purpose is to evaluate the influence of \emph{subtle} and \emph{indistinctive} gestures on the performance of each of the models. These types of movements are more pronounced in co-speech gestures, as described earlier. Additionally, this gives more insights into the performance of each model across different gesture domains.


\subsection{Experimental Settings}
\label{subsec:experimental_settings}
The training parameters of each experiment were selected using grid search and are listed in the next sections. We run each of the models under similar conditions. The hardware specifications used for training and testing the models are as follows: 1) Ubuntu 18.04.5 LTS operating system; 2) Intel Core i7-4930K 3.40 GHz with six cores; 3) 8 GB of RAM; 4) NVIDIA GeForce GTX 1080 graphics card with 8 GB of memory. The performance of each model is evaluated using accuracy, F1-score, and training time metrics. We report the average performance of each model over five trials. In each trial, we repeat the steps of training and testing. Further classification behavior analysis is done by visualizing and discussing the confusion matrices.

\subsection{GRIT Experiment}
\label{subsec:grit_experiment}
The search space and optimal hyperparameters of the GRIT experiment are listed in Table~\ref{table:grit_hyperparameters}. In this experiment, the resulting optimal values are identical, which we explain by the similarity in architecture and training procedure across the models. We use the same data split ratio for each trained model to conduct a fair comparison. To avoid data imbalances, we use stratified sampling in terms of class labels. Similar to the original CNNLSTM proposal~\cite{tsironi16}, we use cross-validation. However, we opt to use a 3-fold split, meaning that approximately 33\% of the data is held out for testing. 
The results of the experiment are summarized in Table~\ref{table:grit_results}. 

Our \emph{Snapture} approach achieves slightly superior results compared to the CNNLSTM in terms of accuracy and F1-score. The scores across the \emph{Snapture} and \emph{Snapture\textsubscript{thold}} variations are similar. 
The three models have a slight deviation across the five trials. We explain the marginal accuracy boost by three factors. First, GRIT robot commands are designed to have a unique motion path, as motivated earlier. Therefore, the CNNLSTM model is sufficient to provide good performance due to its powerful movement learning capabilities. Second, due to the \emph{repeating pattern} gestures, the majority of the GRIT movements do not have sufficient pause for capturing a \emph{snapshot} (approximately 44\%) according to our threshold definition. Combined with the small dataset size, our model may not have seen enough training data to learn the unique characteristics of handshapes. Third, since only approximately 44\% of GRIT samples include a motion at the peak beneath the defined threshold, the \emph{Snapture\textsubscript{thold}} acts similar to a CNNLSTM model in 56\% of the cases. Therefore, it is not able to contribute to a noticeable accuracy increase.

\begin{figure}
    \centering
    \includegraphics[width=0.75\textwidth]{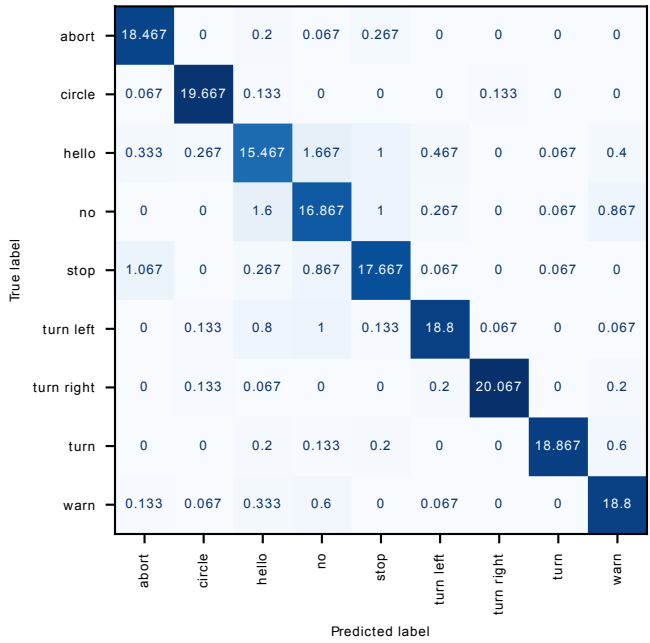}
\caption{The confusion matrix of the average case for the CNNLSTM on the GRIT dataset. The confusion is pronounced between the classes ``hello'' and ``no'', ``hello'' and ``stop'', ``no'' and ``stop''.
}
\label{fig:grit_cnnlstm_confusion}
\end{figure}

\begin{figure}
    \centering
    \includegraphics[width=0.75\textwidth]{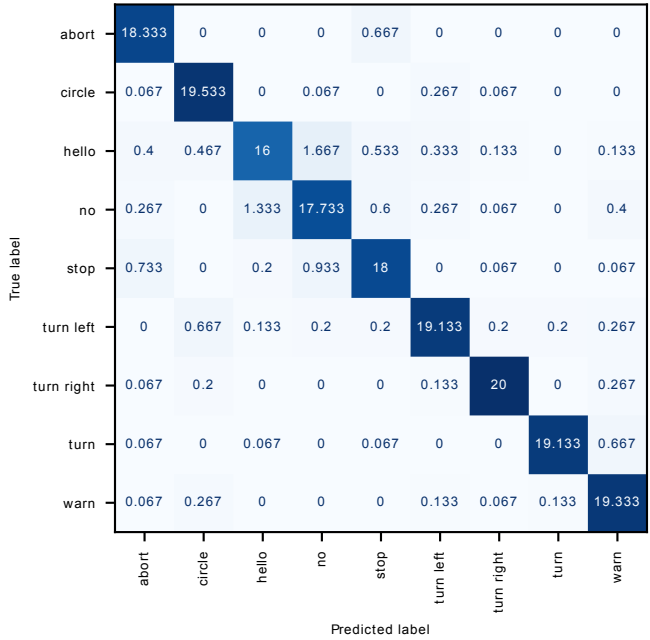}
\caption{The confusion matrix of the average case for \emph{Snapture} on the GRIT dataset. The confusion is less pronounced between the classes ``hello'' and ``no'', ``hello'' and ``stop'', ``no'' and ``stop''. However, the performance is still negatively influenced by the false classification of some ``hello'' samples as ``no''.
}
\label{fig:grit_snapture_confusion}
\end{figure}

\begin{figure}
    \centering
    \includegraphics[width=0.75\textwidth]{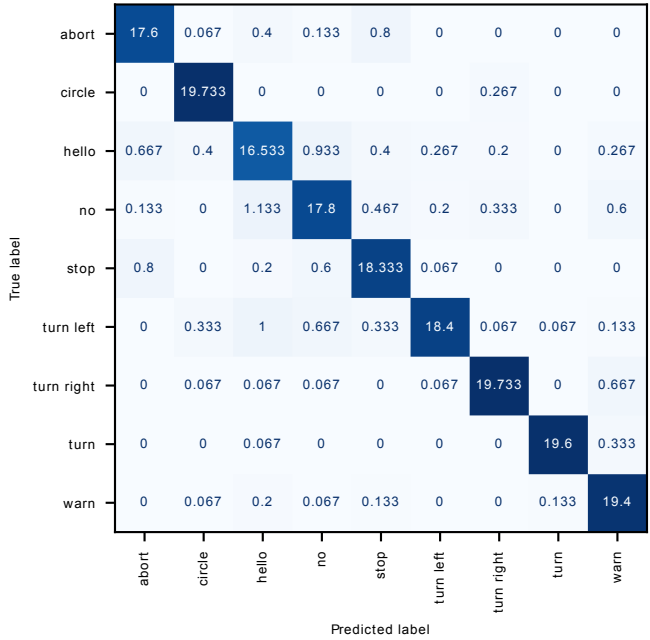}
\caption{The confusion matrix of the average case for \emph{Snapture\textsubscript{thold}} on the GRIT dataset. Less confusion can be observed concerning class ``circle'', which confirms that the static channel should be disabled for such \emph{repeating pattern} movements.
}
\label{fig:grit_thold_confusion}
\end{figure}

\begin{figure}
    \centering
    \includegraphics[width=0.75\textwidth]{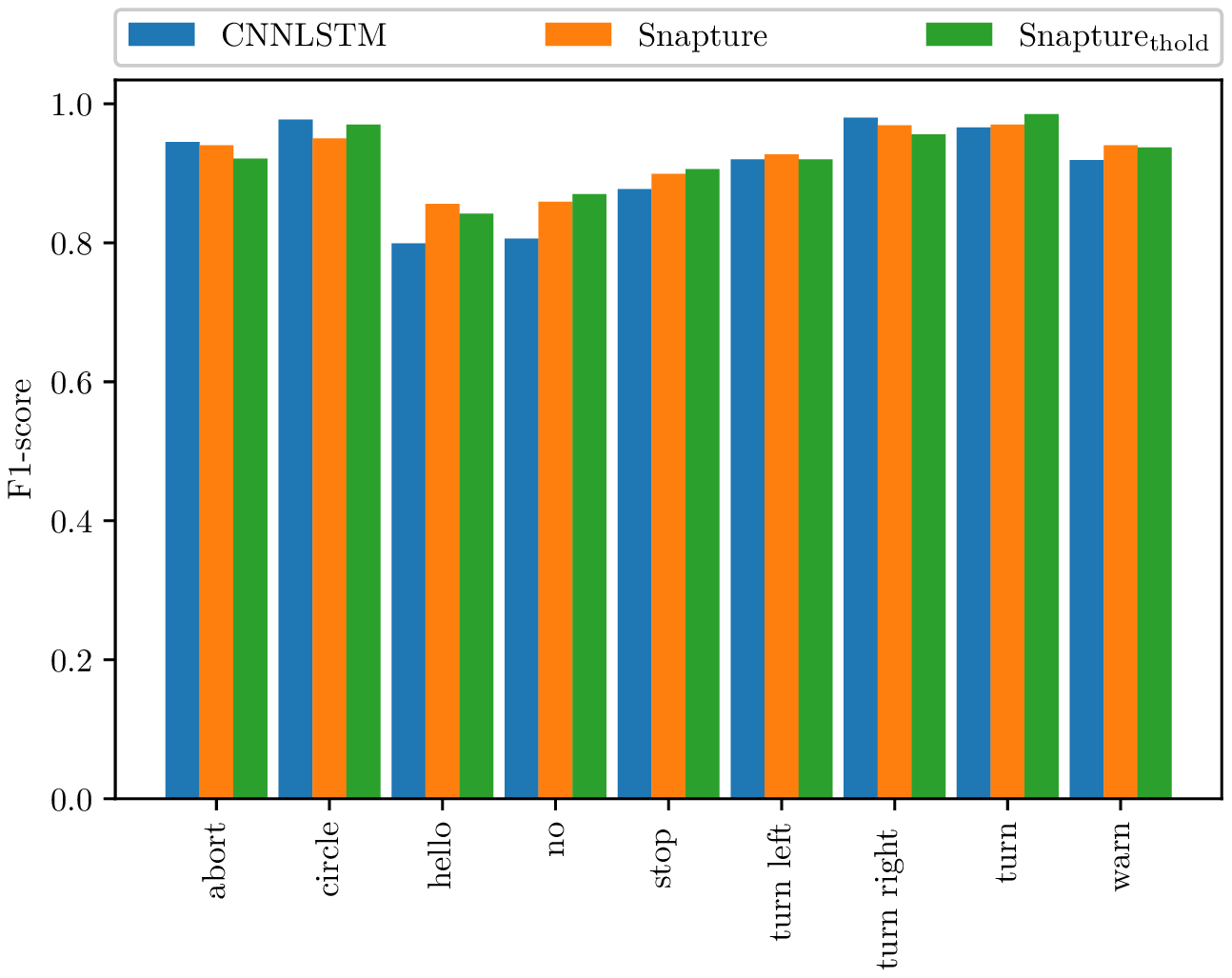}
\caption{A comparison of per-class F1-score values between the different approaches on the GRIT dataset. \emph{Snapture} increases the score on classes ``hello'' and ``no'', while the performance across the remaining classes is comparable.
}
\label{fig:grit_bar_chart}
\end{figure}

However, we analyze the results further through the confusion matrix of the average case, i.e., the mean results of five trials (cf.~Fig.\ref{fig:grit_cnnlstm_confusion}). The most confusion in the CNNLSTM model occurs between the classes ``hello'' and ``no'', ``hello'' and ``stop'', ``no'' and ``stop'', and ``stop'' and ``abort''. All of these movements have a similar motion profile but differ in hand shape. Thus, this supports that the \emph{indistinctive} movements negatively influence the performance of CNNLSTM. We explain that by the CNNLSTM's lack of considering the hand details. Similar findings were reported in the work of Tsironi et al.~\cite{tsironi17}. On the other hand, the confusion between these classes is less pronounced in \emph{Snapture} (cf.~Fig.\ref{fig:grit_snapture_confusion}) due to the static channel, which provides the hand pose information. However, the misclassification of ``hello'' samples as ``no'' still negatively impacts the performance of \emph{Snapture}. We observe that some participants perform ``hello'' and ``no'' in a rapid fashion resulting in a \emph{blur} effect and noisy input to the network. Therefore, the \emph{Snapture\textsubscript{thold}} improves the situation (cf.~Fig.\ref{fig:grit_thold_confusion}) by excluding the \emph{snapshot} in case the input is not sufficient for interpreting the hand details. However, dealing with the low resolution of the dataset remains a challenge to any approach while extracting meaningful hand shape and finger arrangement. 
On the other hand, \emph{repeating pattern} movements, e.g., ``circle'' yields comparable F1-score values across the three architectures (cf.~Fig.\ref{fig:grit_bar_chart}) due to the distinctive hand movement. However, the number of false positives and true negatives associated with ``circle'' drops noticeably in \emph{Snapture\textsubscript{thold}}, further emphasizing that the static channel is indeed counterproductive for such movements.

On a different note, the confusion between the classes ``no'' and ``stop'' is less pronounced in \emph{Snapture} and \emph{Snapture\textsubscript{thold}} compared to CNNLSTM. Despite the dissimilarity between the two classes, some subjects tend to perform ``no'' with a slight left and right hand movement around the wrist, making it very similar to ``stop'' in terms of arm movement (raised and direct towards the camera). As motivated earlier, due to the loss in hand details, the CNNLSTM struggles with this sort of \emph{implicit} hand movements. Therefore, our approach improves the performance by integrating the hand shape and finger arrangements.

\begin{table}[]
\centering
\caption{The search space and optimal hyperparameter values (in bold) of each model in the \emph{GRIT} experiment.}
\label{table:grit_hyperparameters}
\begin{tabular}{lll}
\hline
\textbf{Hyperparameter}   & \textbf{CNNLSTM}          & \textbf{Snapture$^{*}$}         \\ \hline
\textbf{Learning rate}    & {[}0.01, \textbf{0.001}, 0.0001{]} & {[}0.01, \textbf{0.001}, 0.0001{]} \\
\textbf{Number of epochs} & {[}10, 20, \textbf{40} {]}         & {[}10, 20, \textbf{40} {]}         \\
\textbf{Mini-batch size}  & {[}16, 32, \textbf{64}, 128{]}     & {[}16, 32, \textbf{64}, 128{]}     \\
\textbf{Optimizer}        & {[}\textbf{Adam}, SGD{]}           & {[}\textbf{Adam}, SGD{]}           \\ \hline
\multicolumn{3}{l}{$^{*}$\footnotesize{\textbf{Similar for \emph{Snapture\textsubscript{thold}}.}}} \\
\end{tabular}
\end{table}

\begin{table}[]
\centering
\caption{The results of the \emph{GRIT} experiment under the described settings. The reported metrics represent the mean of five trials, while the values in parentheses correspond to the standard deviation. The superior accuracy and F1-score values are in bold.}
\label{table:grit_results}
\begin{tabular}{llll}
\hline
\textbf{Model}    & \textbf{CNNLSTM} & \textbf{Snapture} & \textbf{Snapture\textsubscript{thold}} \\ \hline
\textbf{Accuracy} & 0.91 (0.012)     & 0.924 (0.006)     & \textbf{0.926} (0.008)           \\
\textbf{F1-score} & 0.913 (0.012)    & \textbf{0.927} (0.005)     & 0.913 (0.012)           \\
\textbf{Time$^{*}$}     & 140.612 (0.255)  & 170.012 (1.027)   & 125.156 (1.117)         \\ \hline
\multicolumn{3}{l}{$^{*}$In seconds.} \\
\end{tabular}
\end{table}

\subsection{Montalbano Experiment}
\label{subsec:montalbano_experiment}
Similar to the previous experiment, we report the hyperparameters in Table~\ref{table:montalbano_hyperparameters}. Due to the more considerable number of class labels, the search space is extended compared to the GRIT experiment. Since the dataset is part of the \emph{ChaLearn Looking at People} challenge, it is already split into training and test datasets, with each set containing unique subjects. To avoid any influence triggered by subject variability, we implement our own split with data from all participants. We follow this approach since we focus on comparing the classification behavior of the different models rather than comparing them to the benchmark. Furthermore, we increase the size of the test set. Our split consists of 70\% and 30\% of randomly selected data for training and testing, respectively. Stratified sampling is utilized for an approximately uniform distribution of class labels across the sets.

Our \emph{Snapture} approach scores superior accuracy and F1-score compared to CNNLSTM. Also, the \emph{Snapture\textsubscript{thold}} improves the results even further (cf. Table~\ref{table:montalbano_results}). However, we observe a noticeable time increase in \emph{Snapture\textsubscript{thold}}. We explain that the additional check for each sample to identify where it lies in comparison to the defined threshold. Approximately, 70\% of the Montalbano data contain a sufficient pause for a \emph{snapshot}. Thus, it gives more insights into the performance of the \emph{Snapture\textsubscript{thold}} approach. By observing per-class performance, \emph{Snapture} achieves superior per-class F1-scores compared to the CNNLSTM with the exception of ``basta'' (both models achieve an identical score). Furthermore, we report a boost in F1-score on all classes with the \emph{Snapture\textsubscript{thold}}. 

In contrast to robot commands, most co-speech gestures in Montalbano have a similar path of motion. Generally, we observe two main categories of movements in the Montalbano dataset: single-handed and two-handed. In single hand movements, the arm is raised above the head level. Most of these gestures have a noticeably similar motion with a distinctive hand and finger arrangement at the peak. Additionally, some single-hand gestures include a delicate hand movement at the peak. Second, two hand movements require synchronization between the two arms. 

\begin{figure}
    \centering
    \includegraphics[width=1\textwidth]{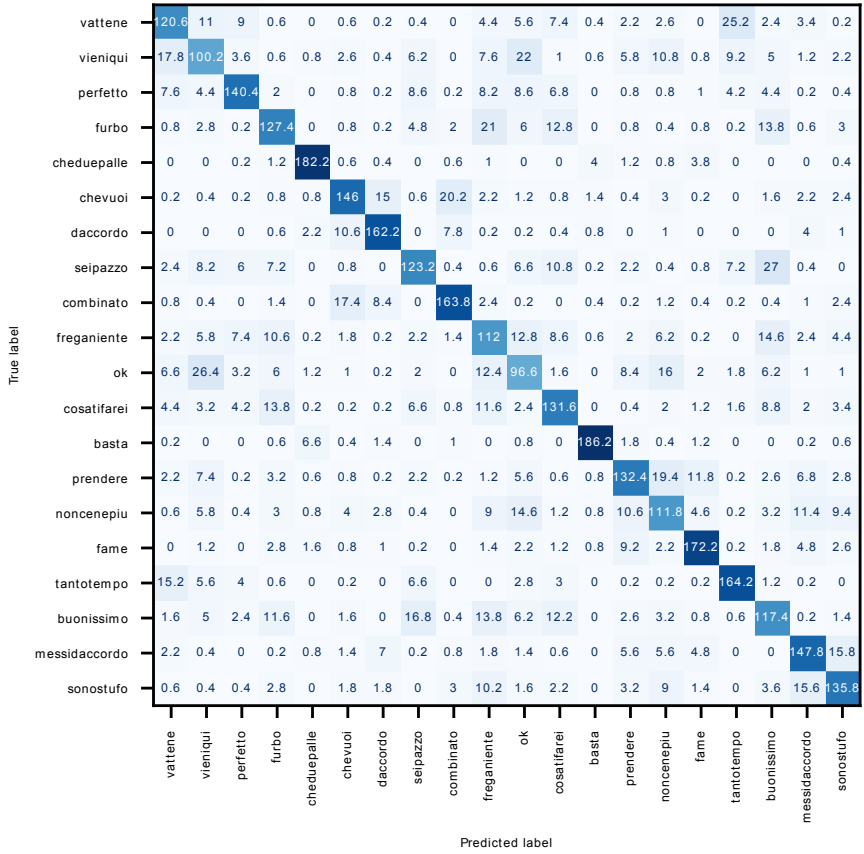}
\caption{The confusion matrix of the average case for CNNLSTM on the Montalbano dataset. The confusion between gesture classes with \emph{indistinctive movement} is pronounced, e.g., ``vattene'', ``vieniqui'', ``perfetto'', and ``tantotempo''.
}
\label{fig:montalbano_cnnlstm_confusion}
\end{figure}

\begin{figure}
    \centering
    \includegraphics[width=1\textwidth]{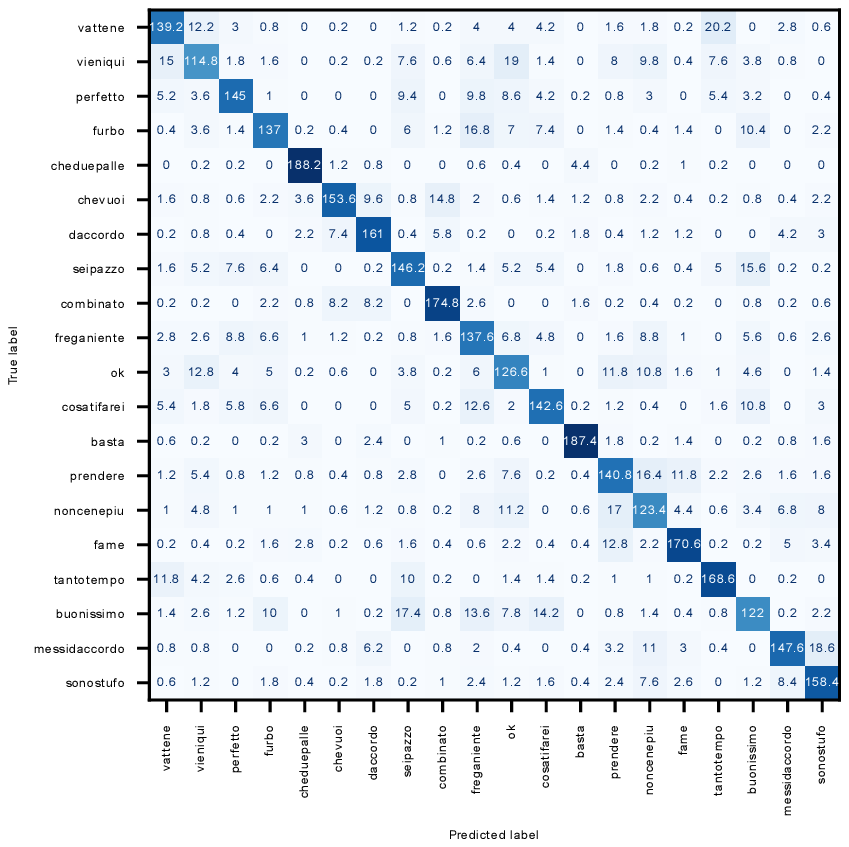}
\caption{The confusion matrix of the average case for \emph{Snapture} on the Montalbano dataset. The confusion concerning gesture classes with \emph{indistinctive} and \emph{implicit} movements, e.g., ``vattene'', ``noncenepiu'' and ``ok'', is less pronounced than the CNNLSTM.
}
\label{fig:montalbano_snapture_confusion}
\end{figure}

\begin{figure}
    \centering
    \includegraphics[width=1\textwidth]{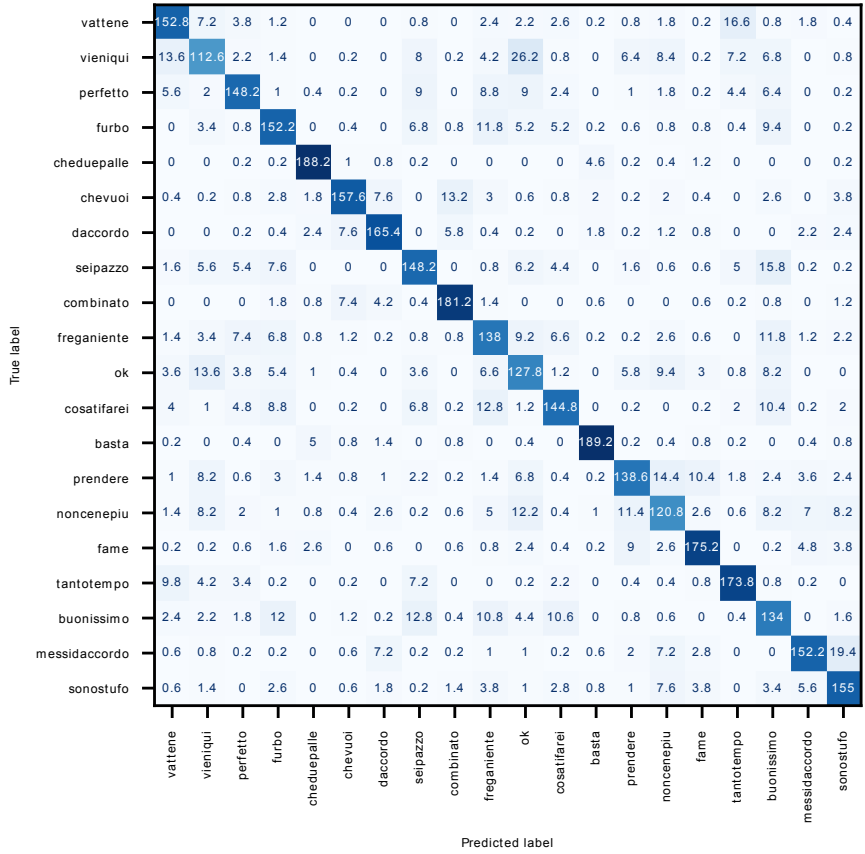}
\caption{The confusion matrix of the average case for \emph{Snapture\textsubscript{thold}} on the Montalbano dataset. The confusion concerning gesture classes ``vattene'', ``furbo'' and ``buonissimo'' is less pronounced than the CNNLSTM.
}
\label{fig:montalbano_thold_confusion}
\end{figure}

\begin{figure}
    \centering
    \includegraphics[width=0.75\textwidth]{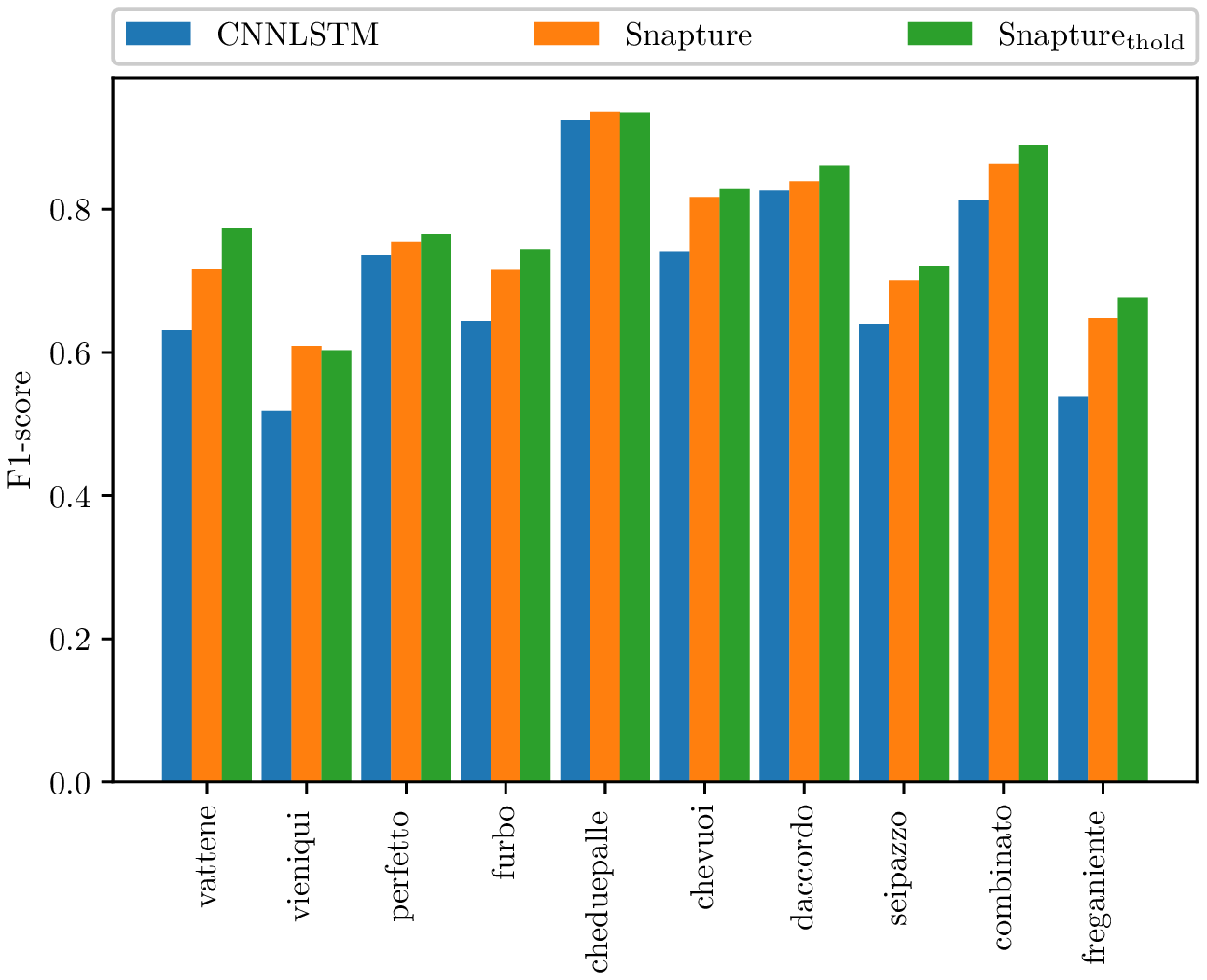}
    \includegraphics[width=0.75\textwidth]{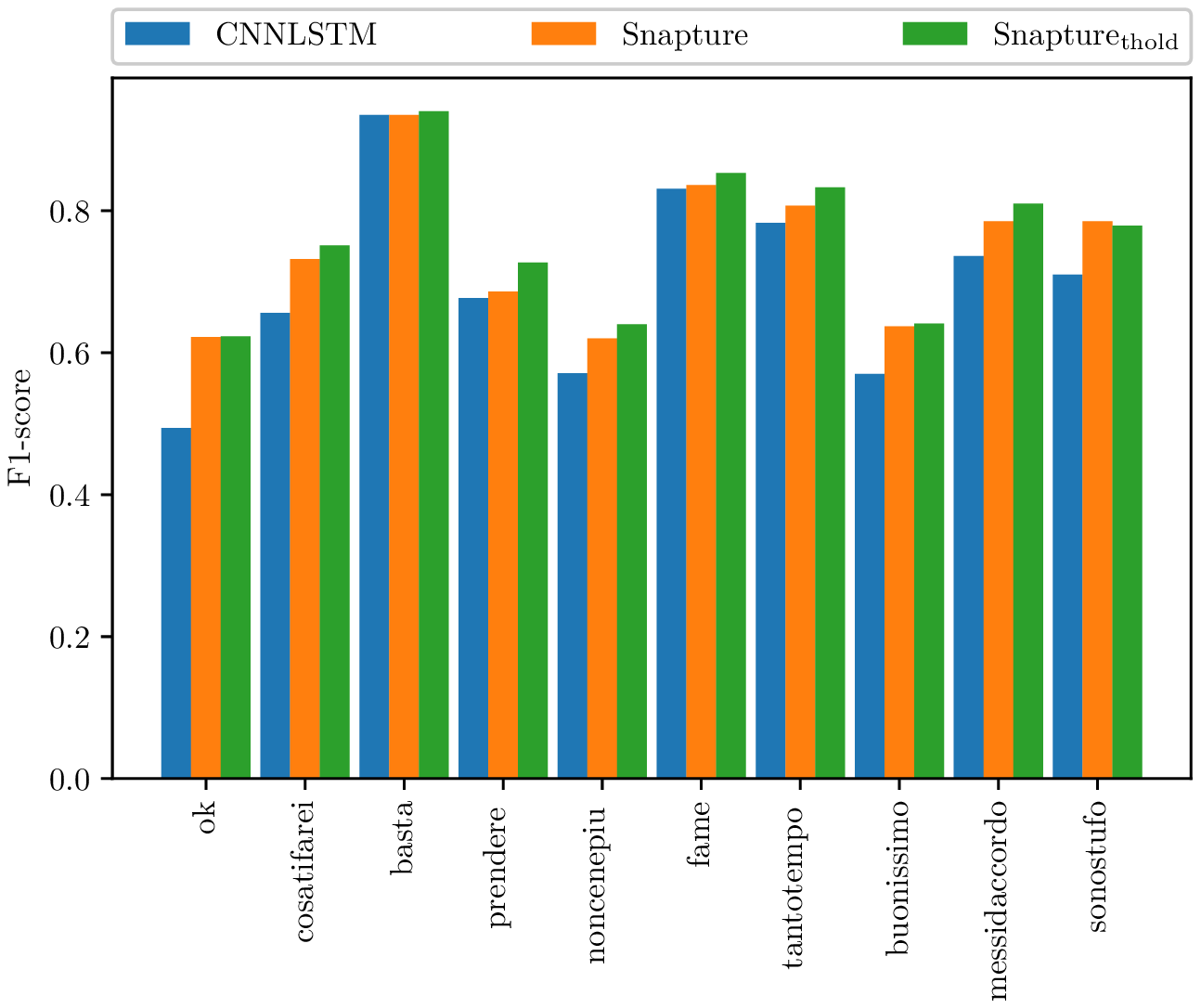}
\caption{A comparison of per-class F1-score values between the different approaches on the Montalbano dataset. \emph{Snapture} improves the score on all classes except ``basta''. The performance of \emph{explicit} arm movements, e.g., ``basta'' and ``cheduepalle'' is comparable across the three models.
}
\label{fig:montalbano_bar_chart}
\end{figure}

\paragraph{Indistinctive Movements:}
In CNNLSTM, multiple observations of classes ``vattene''  are miscalssified as ``vieniqui'', ``perfetto'' or ``tantotempo''  (cf. Fig.~\ref{fig:montalbano_cnnlstm_confusion}). We explain that by the similarity in hand motion. Compared to the CNNLSTM, an addition of $\sim$19 and $\sim$32 samples on average are correctly classified by the \emph{Snapture} and \emph{Snapture\textsubscript{thold}}, respectively (cf. Fig.~\ref{fig:montalbano_snapture_confusion} and Fig.~\ref{fig:montalbano_thold_confusion}). Consequently, we observe F1-score improvements in the respective classes (cf. Fig.~\ref{fig:montalbano_bar_chart}). Furthermore, the CNNLSTM achieves poor F1-score values (below 0.6) on classes  ``vieniqui'' and ``freganiente'', ``ok'', ``noncenepiu'' and ``buonissimo''. Most of the confusion of class ``ok'' is tied to false positives/negatives with one of the said classes. This can be explained by the similarity in their motion. However, the total number of misclassified ``ok'' samples drops in  \emph{Snapture} and \emph{Snapture\textsubscript{thold}} by approximately 30. Therefore, we observe an increase in the F1-score. 
Additionally, \emph{Snapture} and \emph{Snapture\textsubscript{thold}} enhance the F1-score of class ``seipazzo''. Approximately, an additional average of 23 and 25 samples are correctly classified due to less confusion with ``buonissimo''. 

\begin{figure}%
    \captionsetup[subfigure]{labelformat=empty}
    \centering
    \subfloat[\centering vattene]{{\includegraphics[width=3cm]{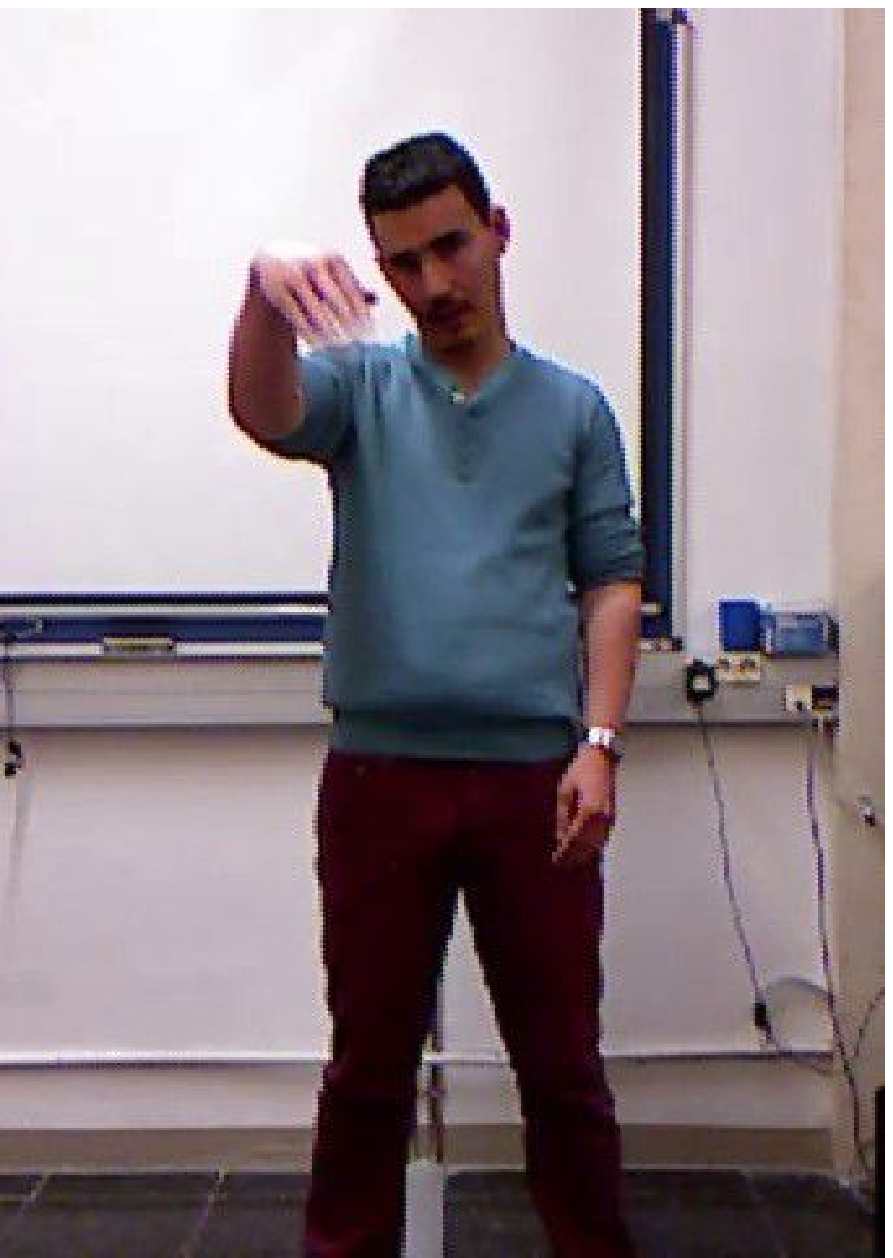} }}%
    \qquad
    \subfloat[\centering vieniqui]{{\includegraphics[width=3cm]{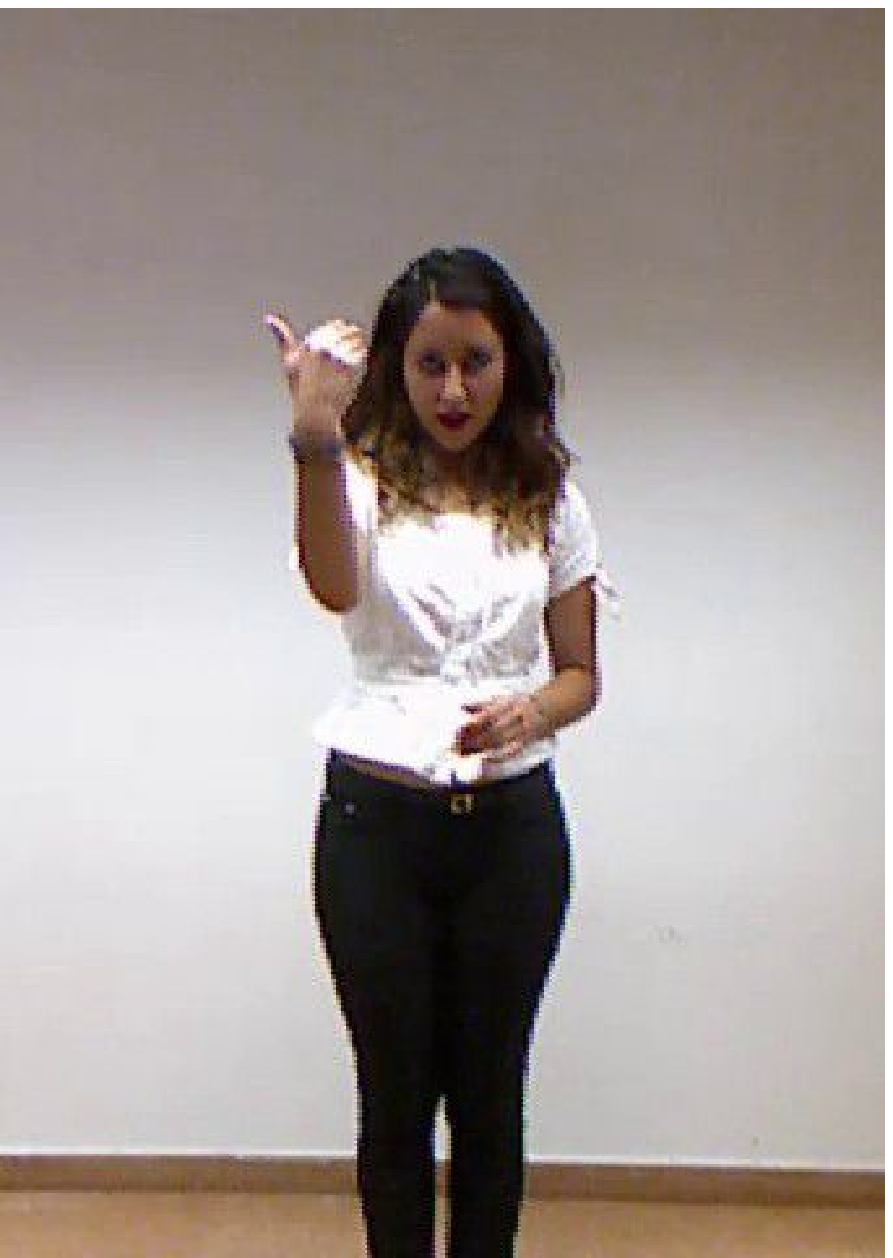} }}%
    \qquad
    \subfloat[\centering tantotempo]{{\includegraphics[width=3cm]{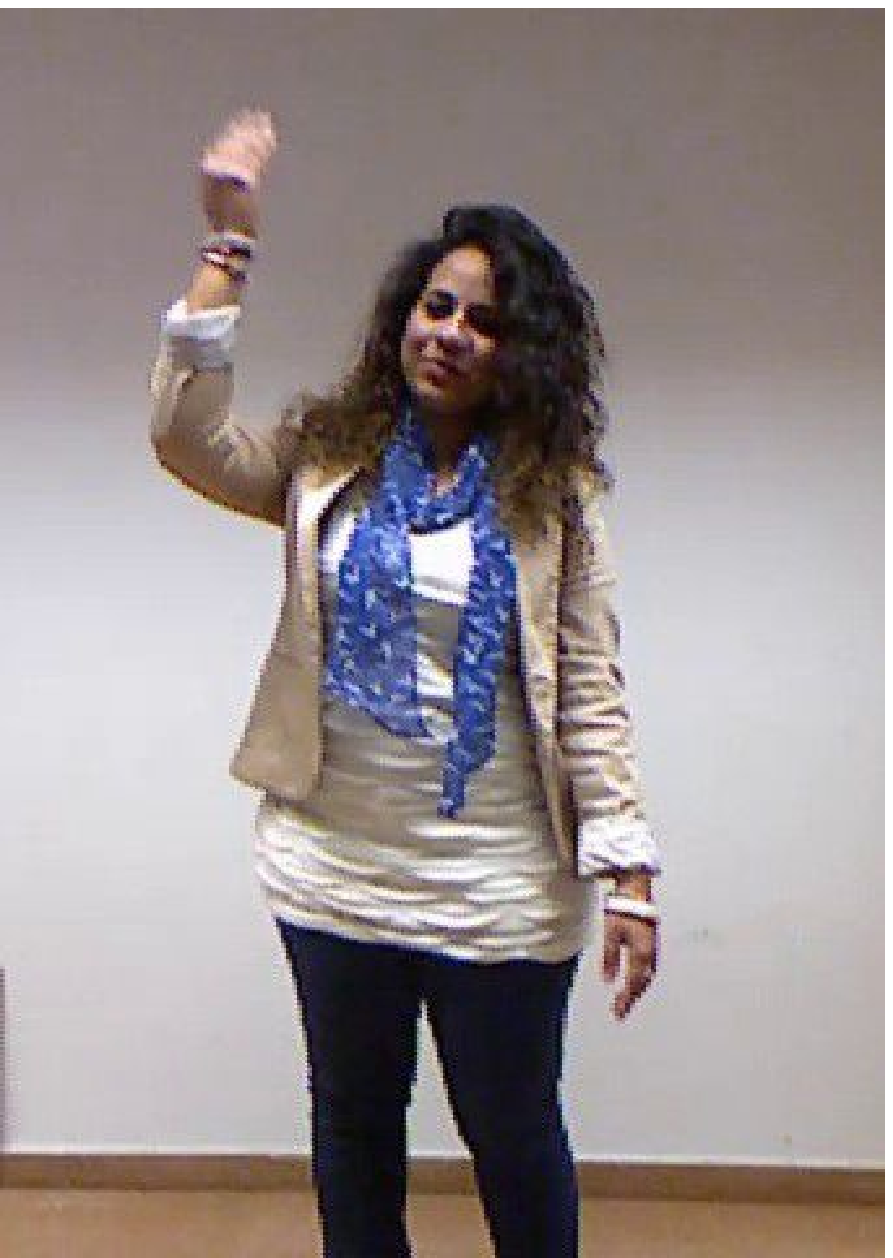} }}%
    \qquad
    \qquad
    \subfloat[\centering vattene (snapshot)]{{\includegraphics[width=3cm]{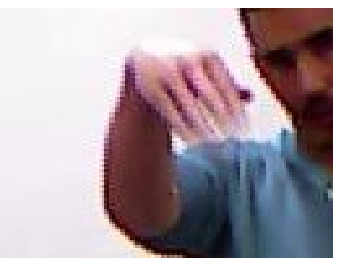} }}%
    \qquad
    \subfloat[\centering vieniqui (snapshot)]{{\includegraphics[width=3cm]{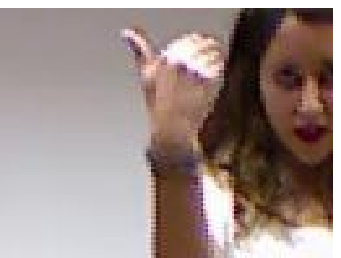} }}%
    \qquad
    \subfloat[\centering tantotempo (snapshot)]{{\includegraphics[width=3cm]{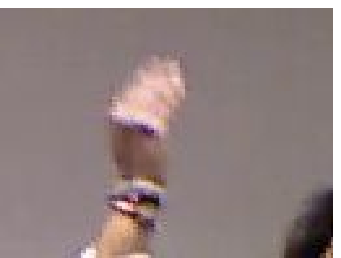} }}%
    \qquad
    \caption{Our \emph{snapshot} extraction takes place using a single frame at the peak. Thus, a challenging scenario to our approach is when gestures that have a similar hand pose during the \emph{stroke} phase. }%
    \label{fig:example_1}%
\end{figure}

On a different note, classes that share both the motion and handshape are challenging for our approach. For example, classes ``vattene'', ``vieniqui'' and ``tantotempo'' use a similar open palm handshape at the peak (cf. Fig.~\ref{fig:example_1}). Therefore, the confusion between such these classes is still noticeable in \emph{Snapture} and \emph{Snapture\textsubscript{thold}} despite the handshape information.

\begin{figure}
\captionsetup[subfigure]{labelformat=empty}
\centering
    \subfloat[\centering furbo]{{\includegraphics[width=0.18\textwidth]{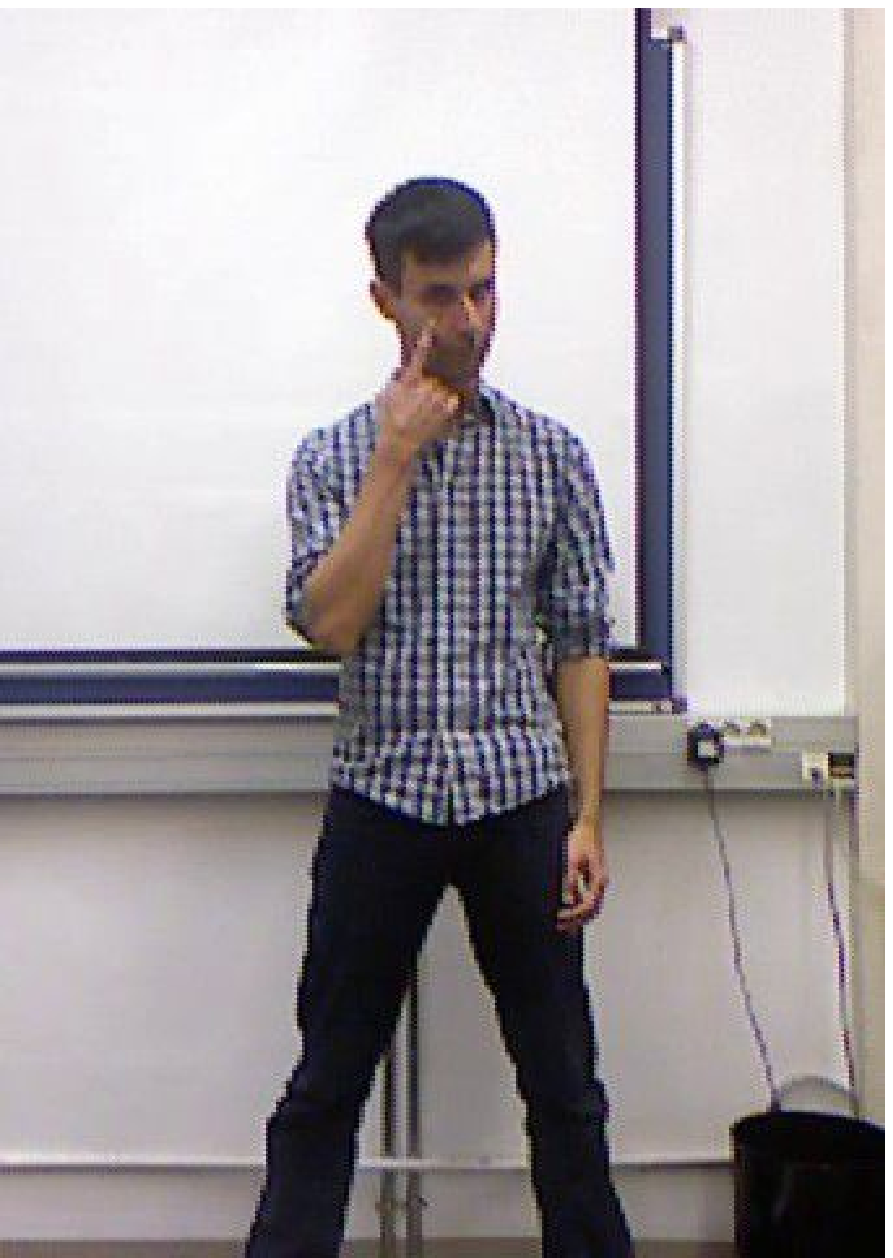} }}%
        \enspace
    \subfloat[\centering buonissimo]{{\includegraphics[width=0.18\textwidth]{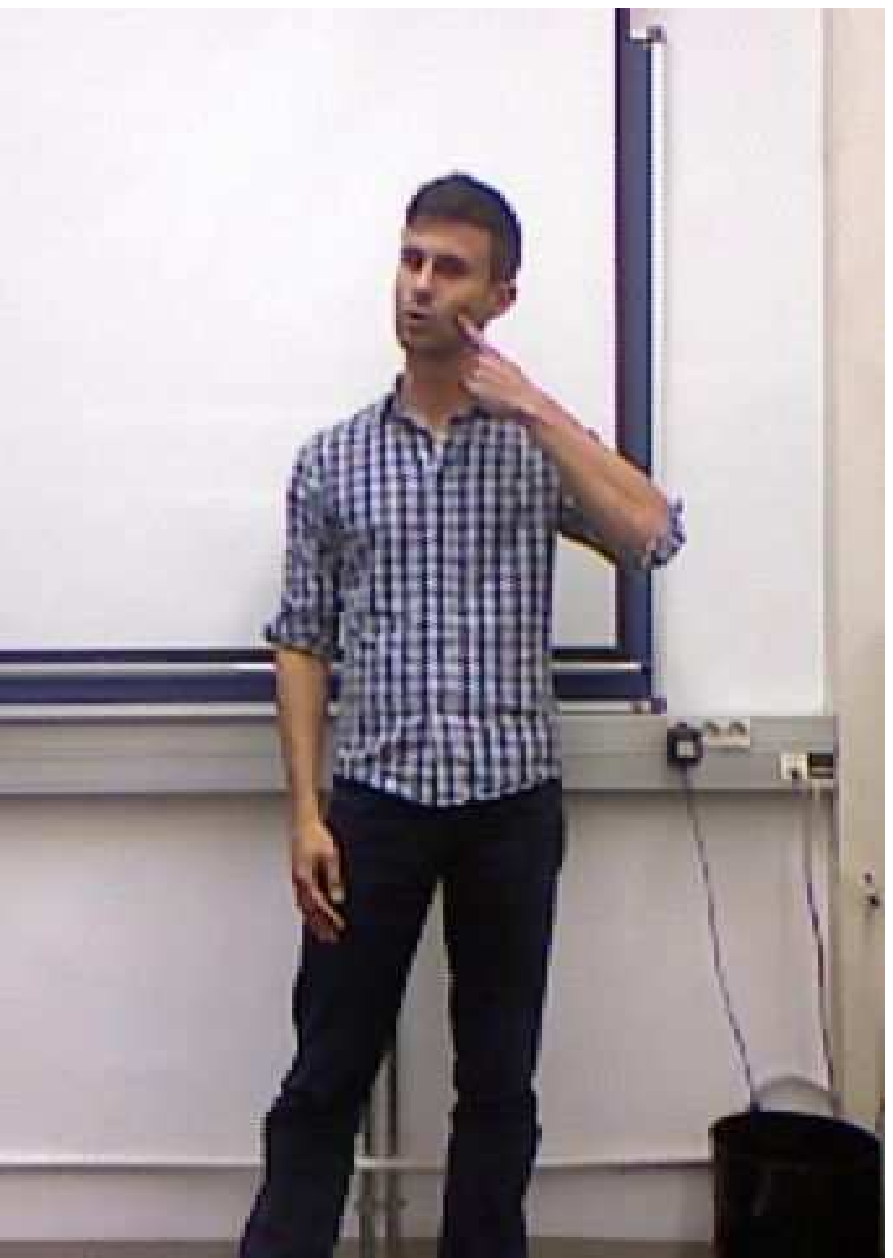} }}%
    \enspace
    \subfloat[\centering freganiente]{{\includegraphics[width=0.18\textwidth]{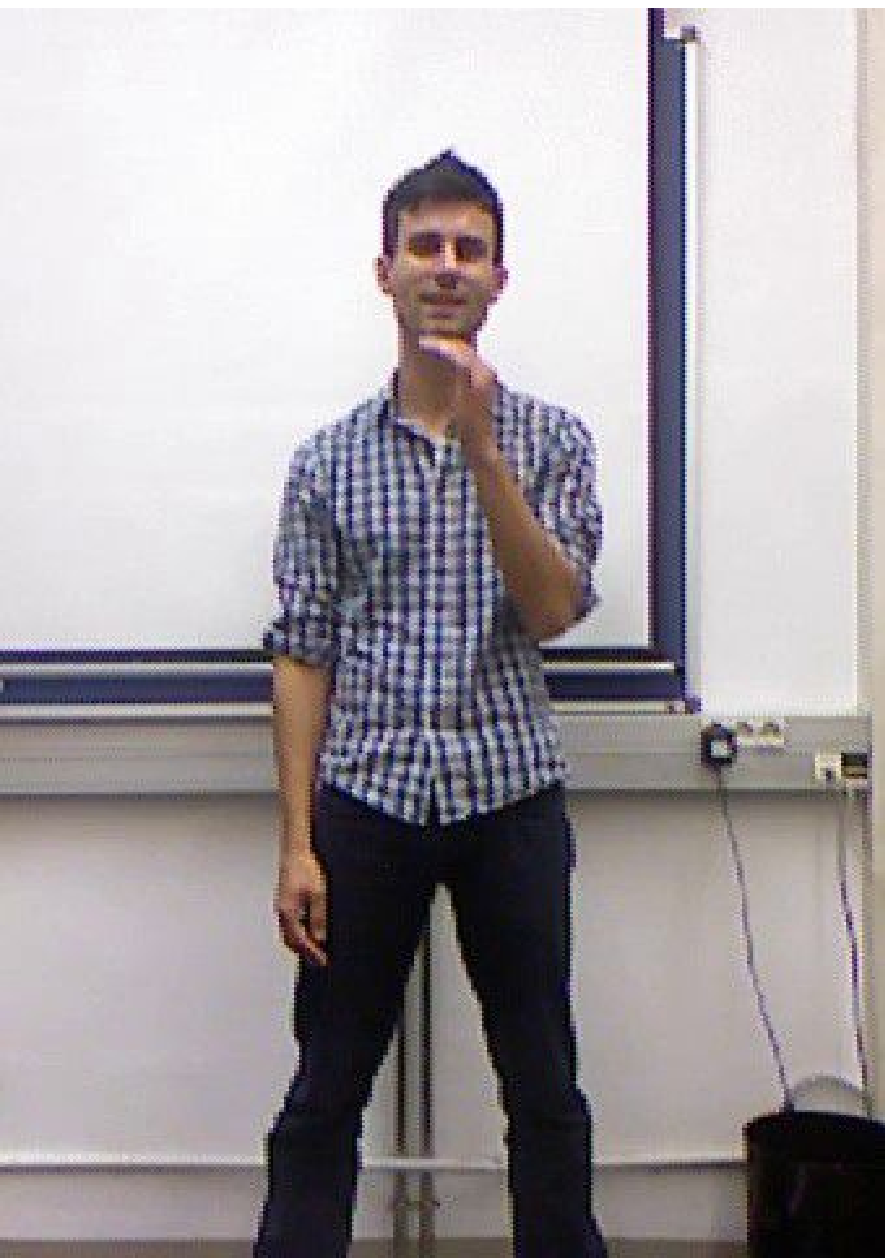} }}%
    \enspace
    \subfloat[\centering cosatifarei]{{\includegraphics[width=0.18\textwidth]{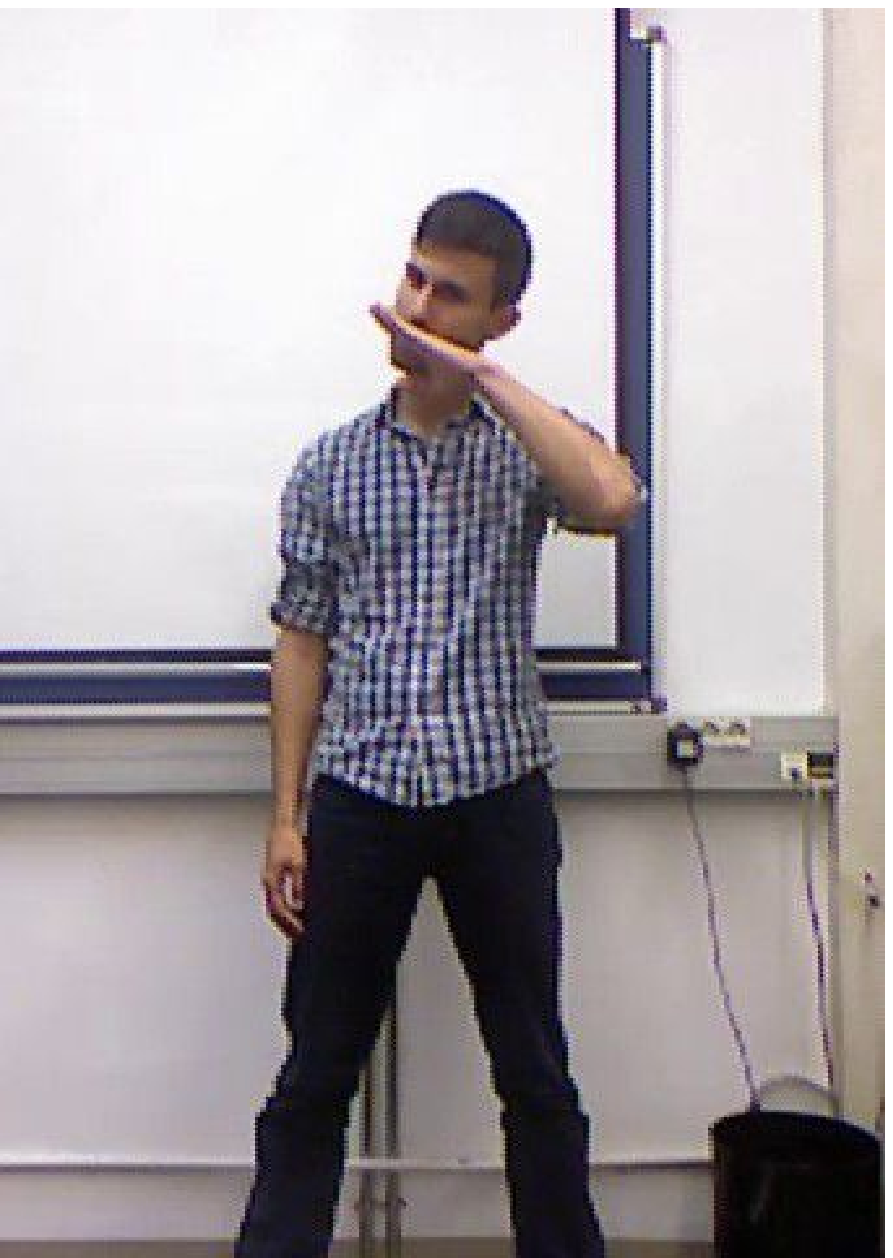} }}%
    \qquad
    \subfloat[\centering furbo (snapshot)]{{\includegraphics[width=0.18\textwidth]{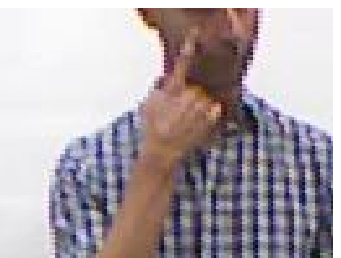} }}%
    \enspace
    \subfloat[\centering buonissimo (snapshot)]{{\includegraphics[width=0.18\textwidth]{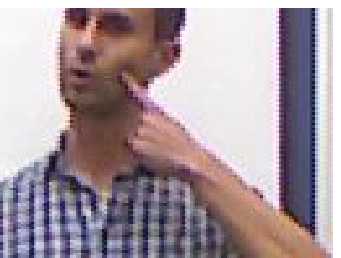} }}%
    \enspace
    \subfloat[\centering freganiente (snapshot)]{{\includegraphics[width=0.18\textwidth]{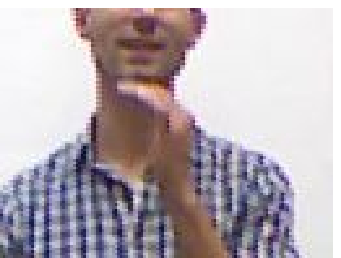} }}%
    \enspace
    \subfloat[\centering cosatifarei (snapshot)]{{\includegraphics[width=0.18\textwidth]{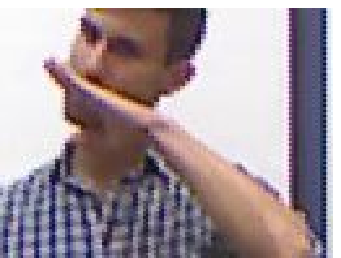} }}%
\caption{Some challenges concerning class ``buonissimo'': a) similarity in hand motion and pose with ``furbo''. Therefore, another modality is required, which is not considered by approach,  b) similarity in hand orientation and light reflection causes misclassifications with ``freganiente'' and ``cosatifarei'' in the worst case of our results. It becomes challenging to interpret the open palm under these conditions.
}
\label{fig:example_2}
\end{figure}

\paragraph{Implicit Movements:}
Besides the motion similarity, some single-hand gestures include a delicate hand movement at the peak. For example, ``sonostufo'' includes a subtle movement of the hand against the chest. Similarly, ``noncenepiu'' and ``buonissimo'' include a rotational motion of the extended index and thumb fingers around the wrist. Due to the pre-processing, i.e., the \emph{differential images} algorithm, these implicit hand details and movements are lost. Consequently, they are not picked up by the CNNLSTM due to the lack of information at the input. However, the confusion related to these classes is noticeably less in \emph{Snapture} and \emph{Snapture\textsubscript{thold}} (cf. Fig.~\ref{fig:montalbano_snapture_confusion} and Fig.~\ref{fig:montalbano_thold_confusion}). On the other hand, the confusion regarding class ``buonissimo'' is only slightly boosted in \emph{Snapture} and \emph{Snapture\textsubscript{thold}}. We explain that by observing that ``buonissimo'' and ``furbo'' are similar in both the motion and handshape, i.e., extended index finger. The difference lies in the position the finger touches the face (under the eyes vs. on the cheek). Efficiently recognizing these gestures requires additional modalities, which we do not consider in our study. However, we will discuss this point later. Moreover, since \emph{snapshot} is captured using one frame at the gesture's peak, it is subject to influence by the corresponding hand orientation and light reflection. Thus, it becomes more challenging to distinguish between an open palm and an extended index finger, especially since the input is in grayscale (cf. Fig.~\ref{fig:example_2}).

\begin{figure}%
    \centering
    \subfloat[]{%
      \begin{minipage}{\linewidth}
          \includegraphics[width=0.18\textwidth]{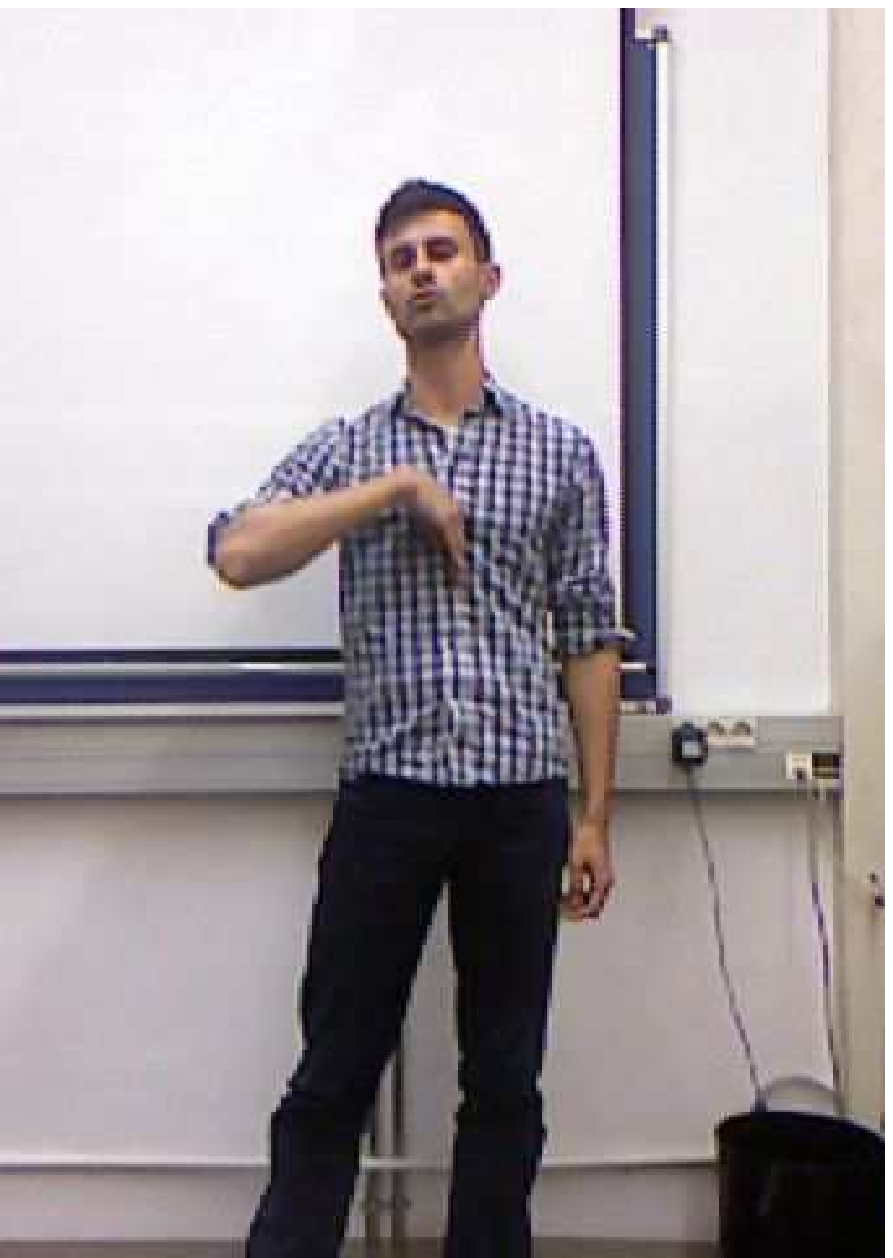}\enspace
          \includegraphics[width=0.18\textwidth]{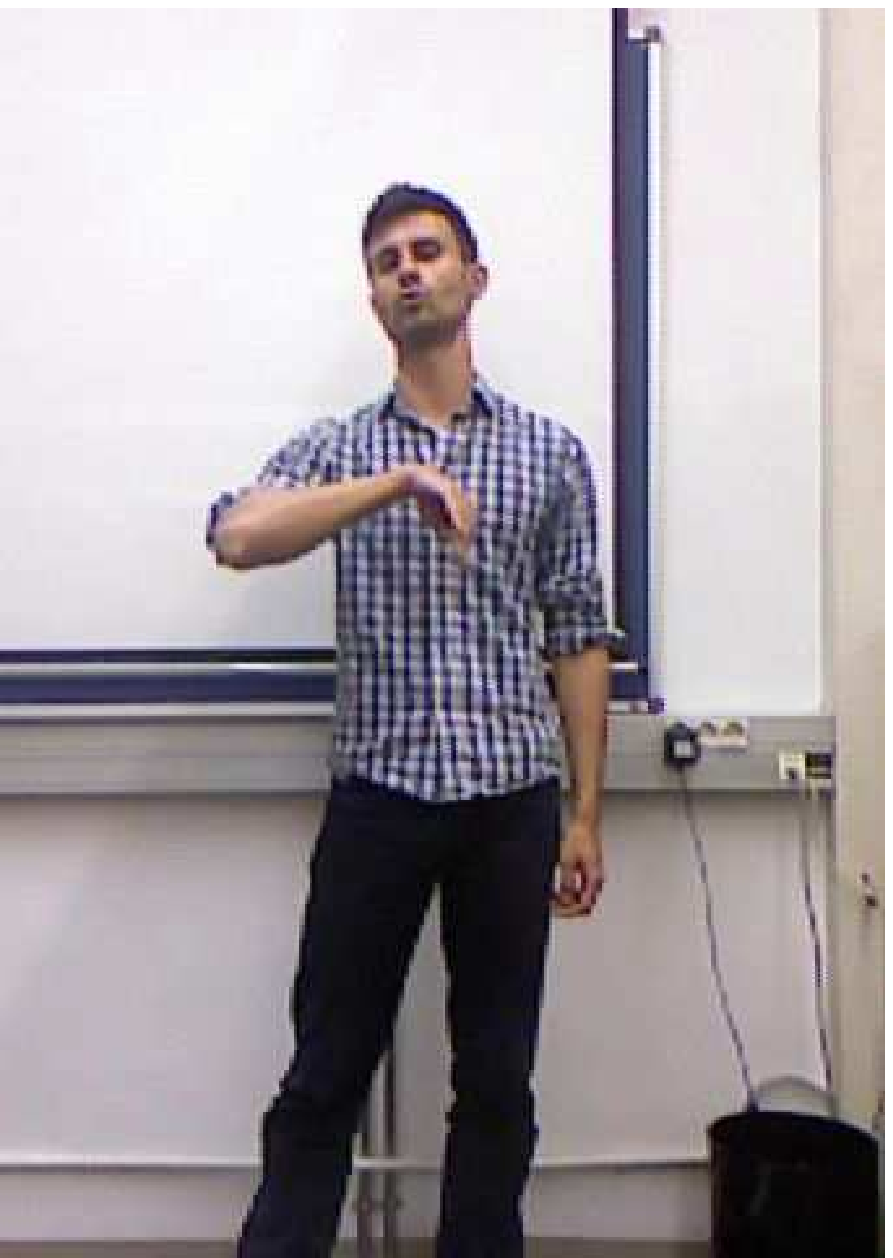}\enspace
          \includegraphics[width=0.18\textwidth]{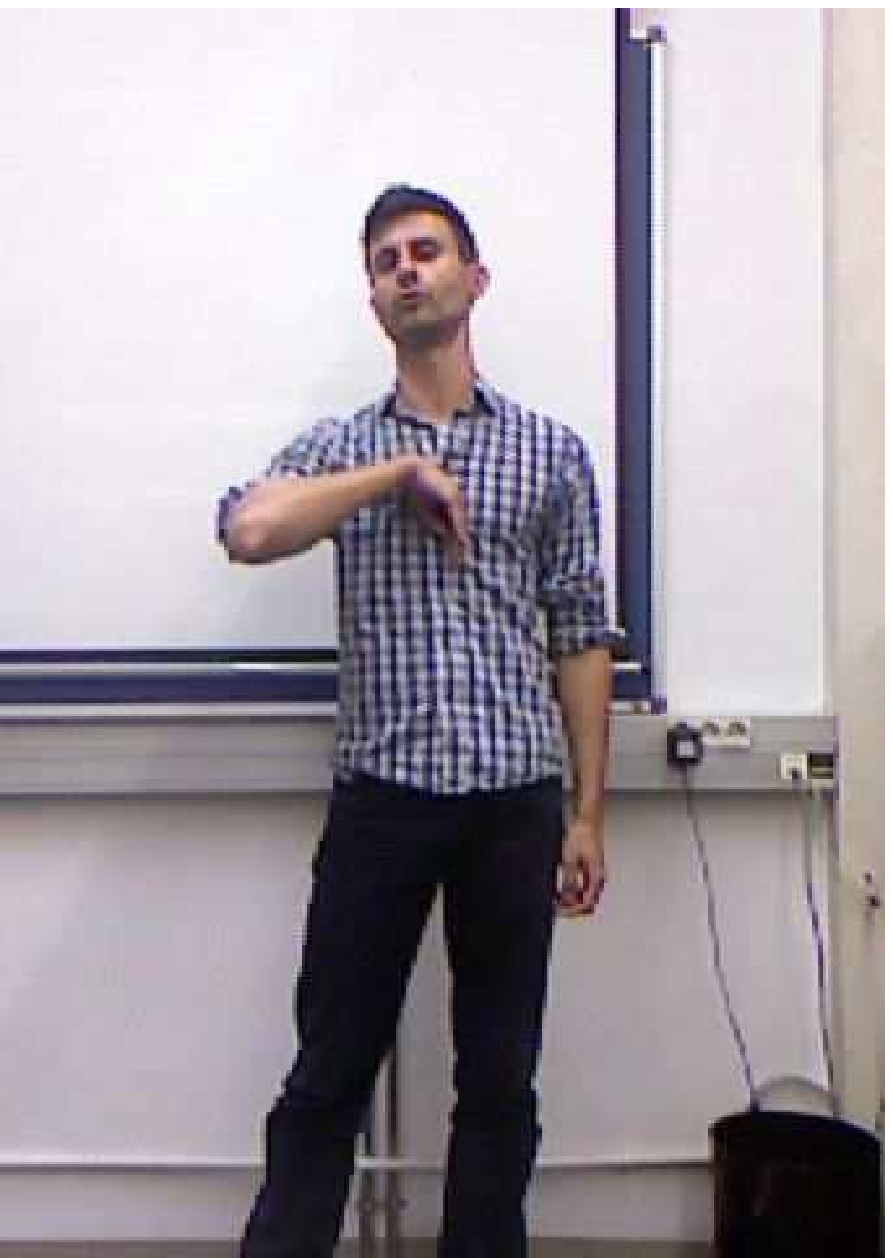}\enspace
          \includegraphics[width=0.18\textwidth]{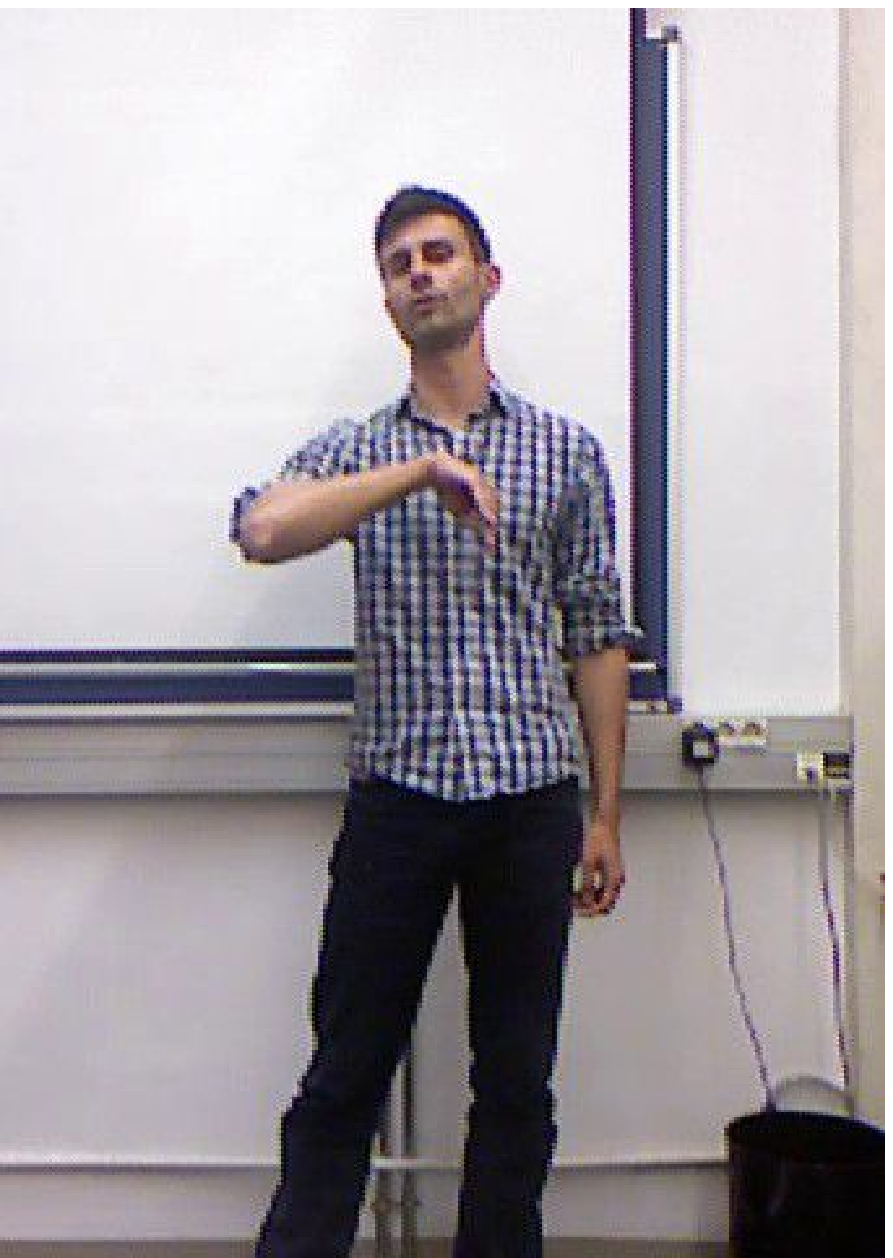}\enspace
          \includegraphics[width=0.18\textwidth]{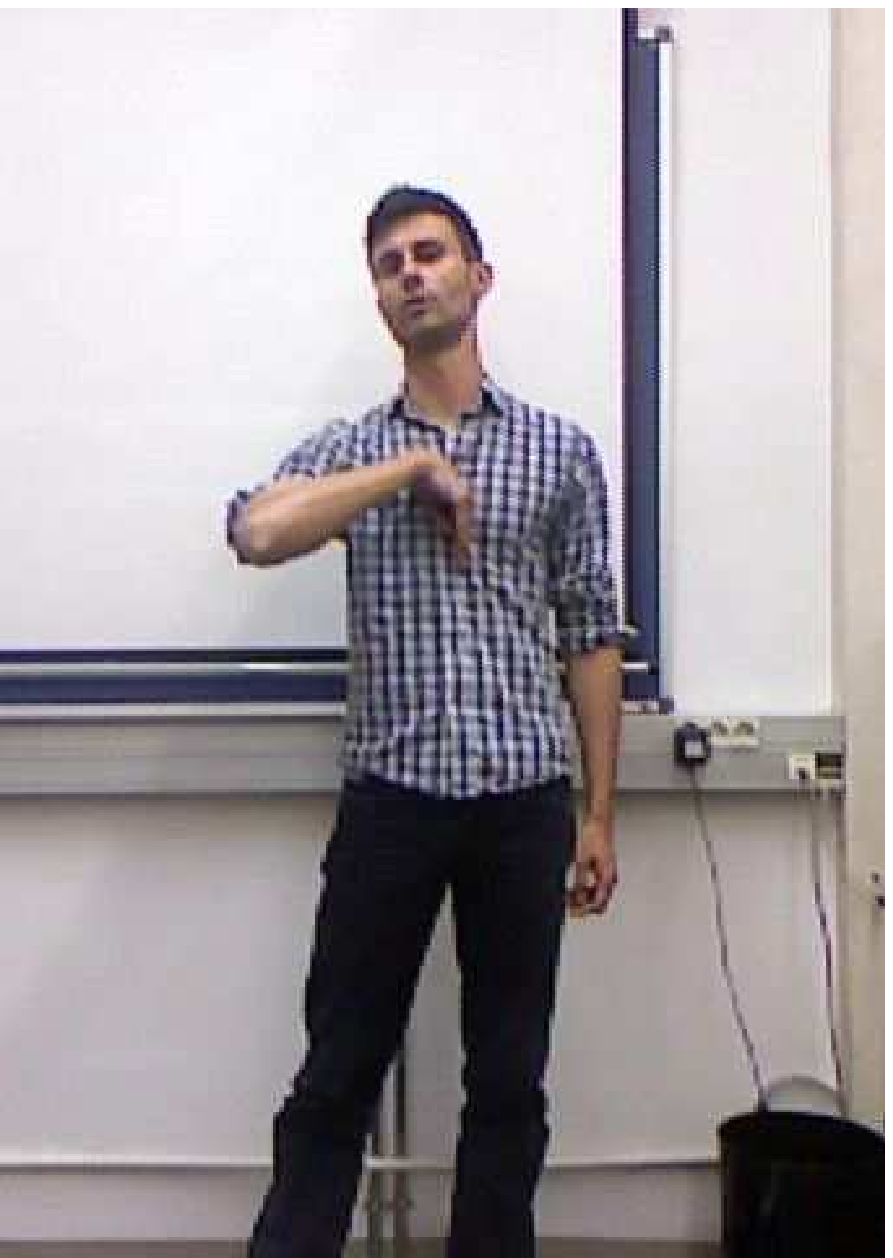}\enspace
      \end{minipage}%
    }\par
    
    \subfloat[]{%
      \begin{minipage}{\linewidth}
          \includegraphics[width=0.18\textwidth]{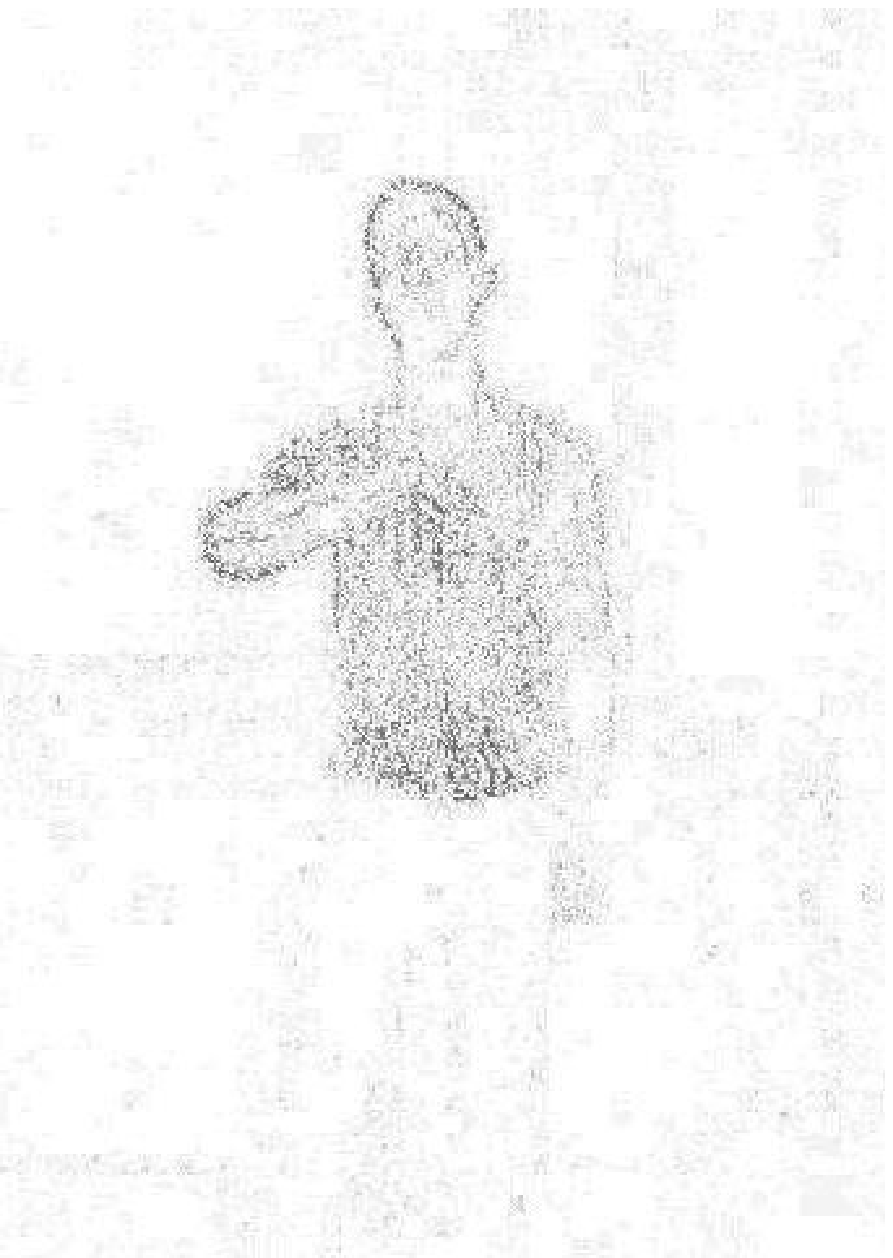}\enspace
          \includegraphics[width=0.18\textwidth]{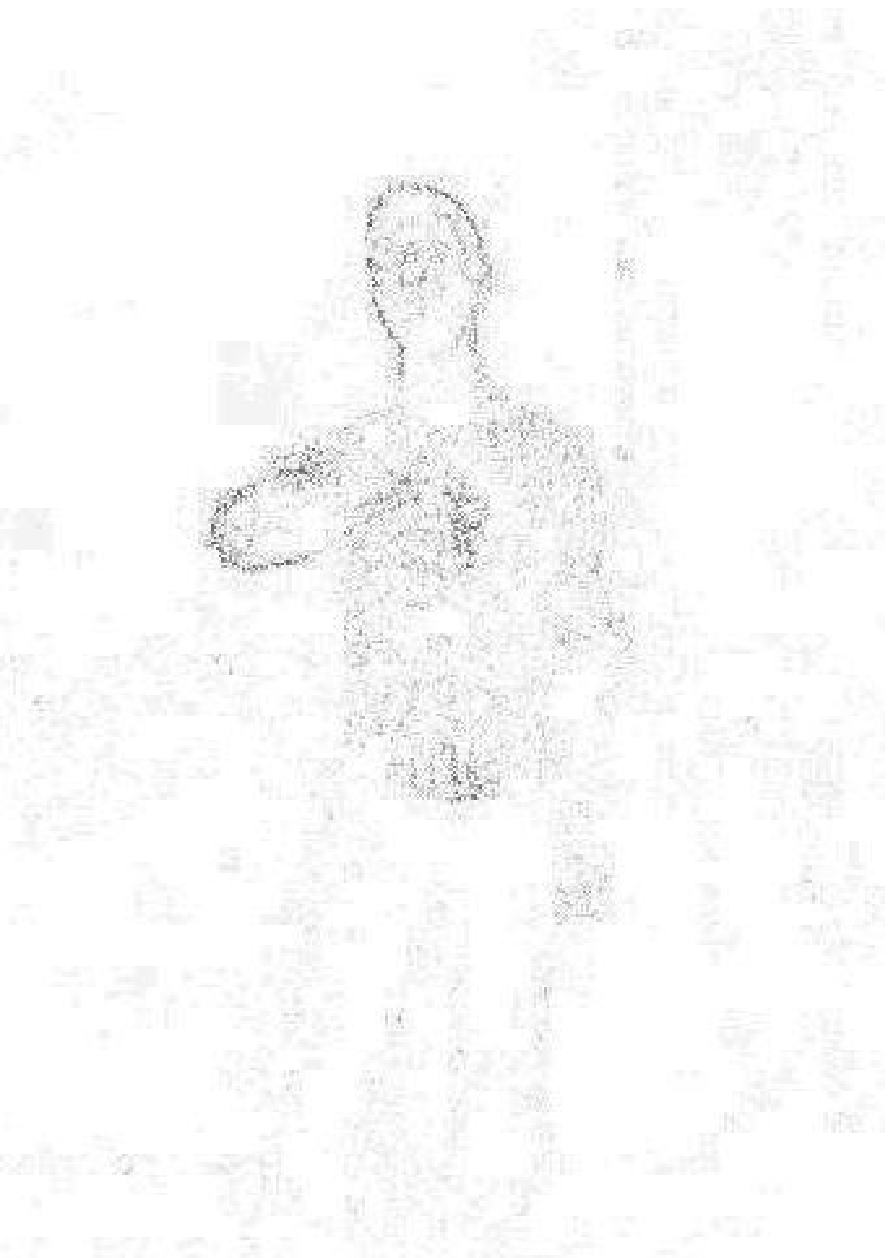}\enspace
          \includegraphics[width=0.18\textwidth]{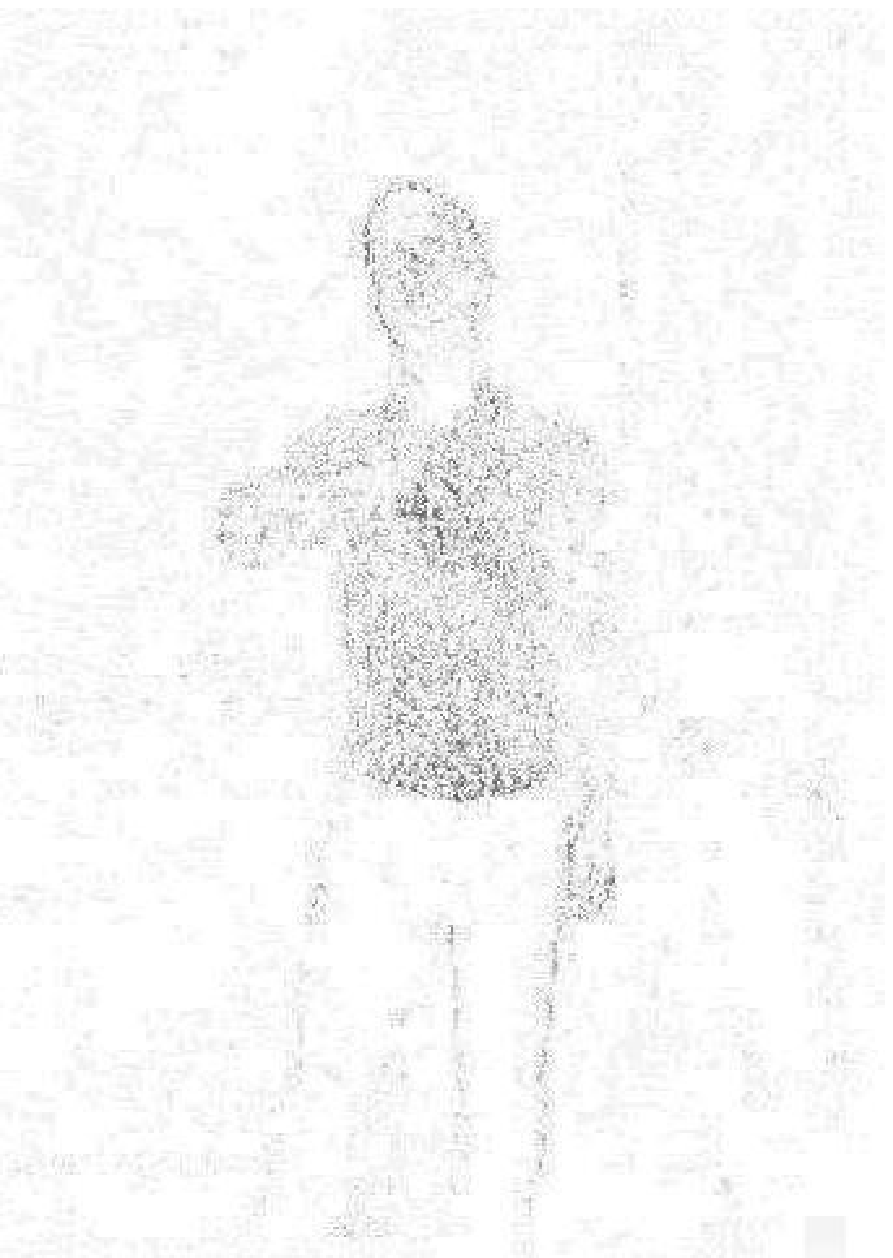}\enspace
          \includegraphics[width=0.18\textwidth]{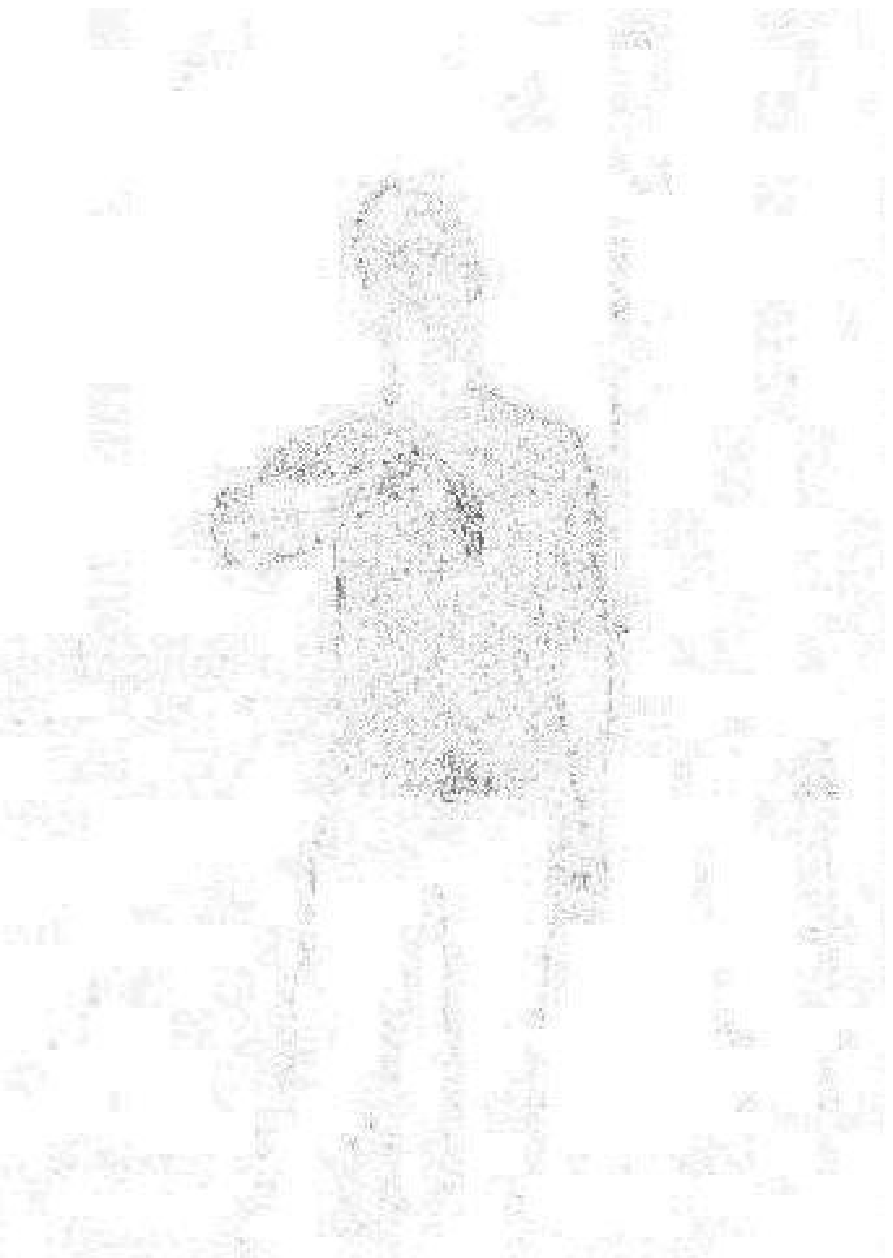}\enspace
          \includegraphics[width=0.18\textwidth]{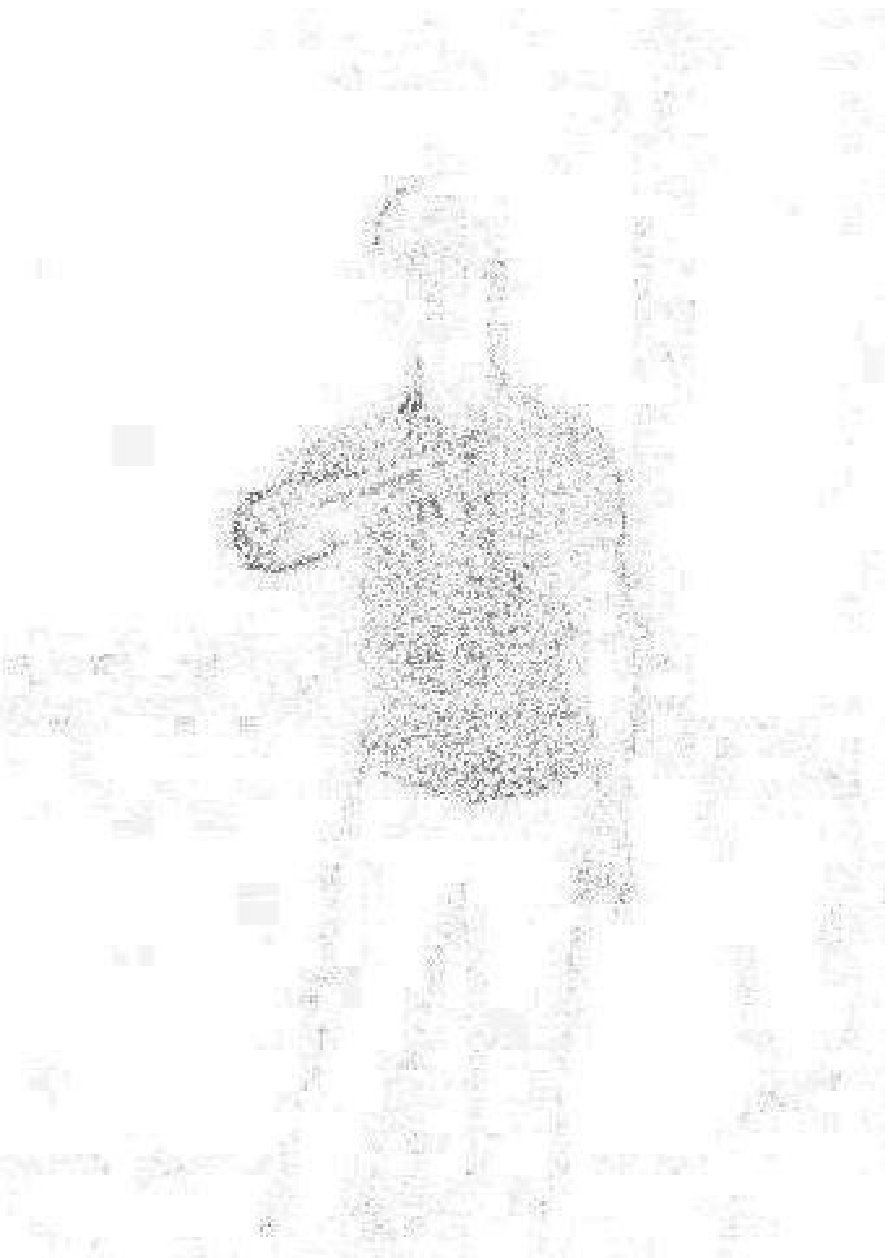}\enspace
      \end{minipage}%
    }\par
    
    \subfloat[]{%
      \begin{minipage}{\linewidth}
          \includegraphics[width=0.18\textwidth]{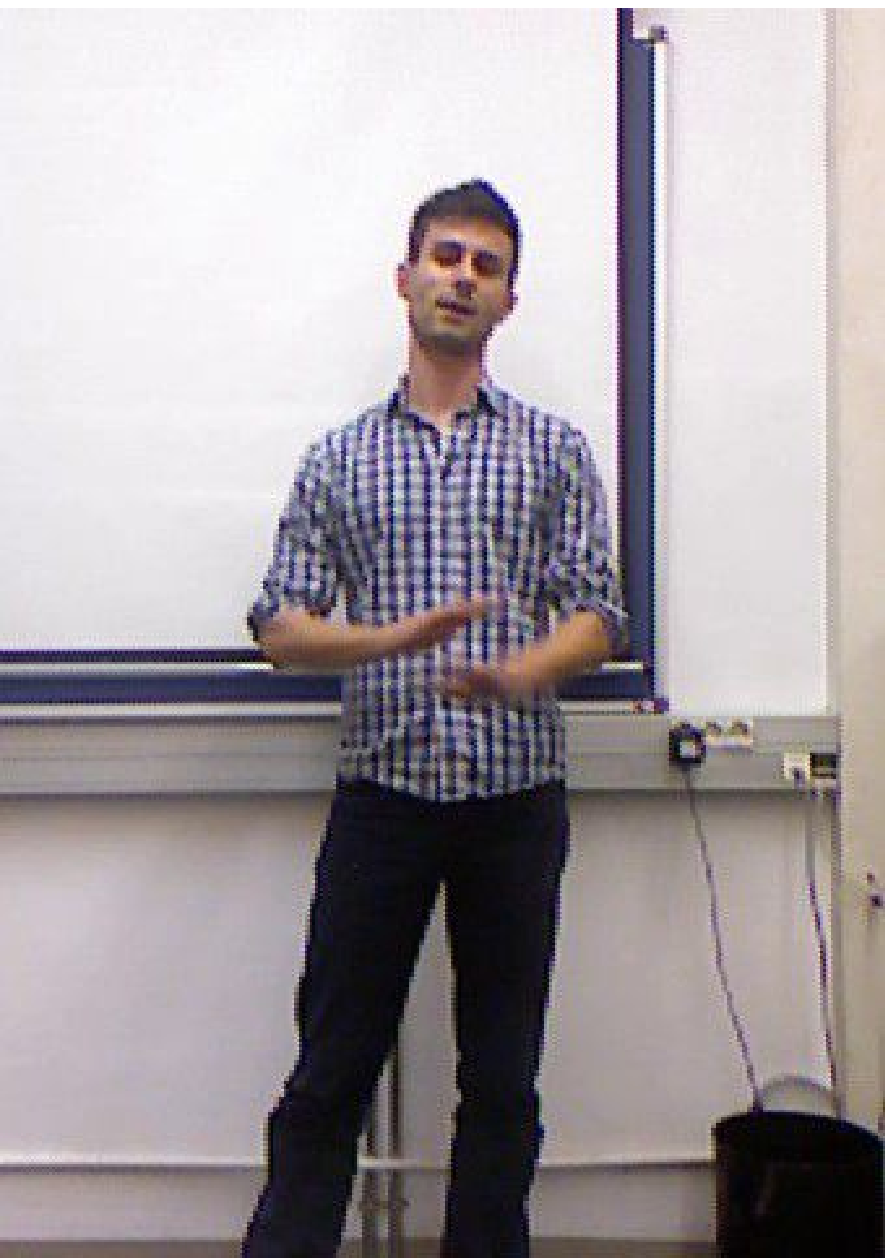}\enspace
          \includegraphics[width=0.18\textwidth]{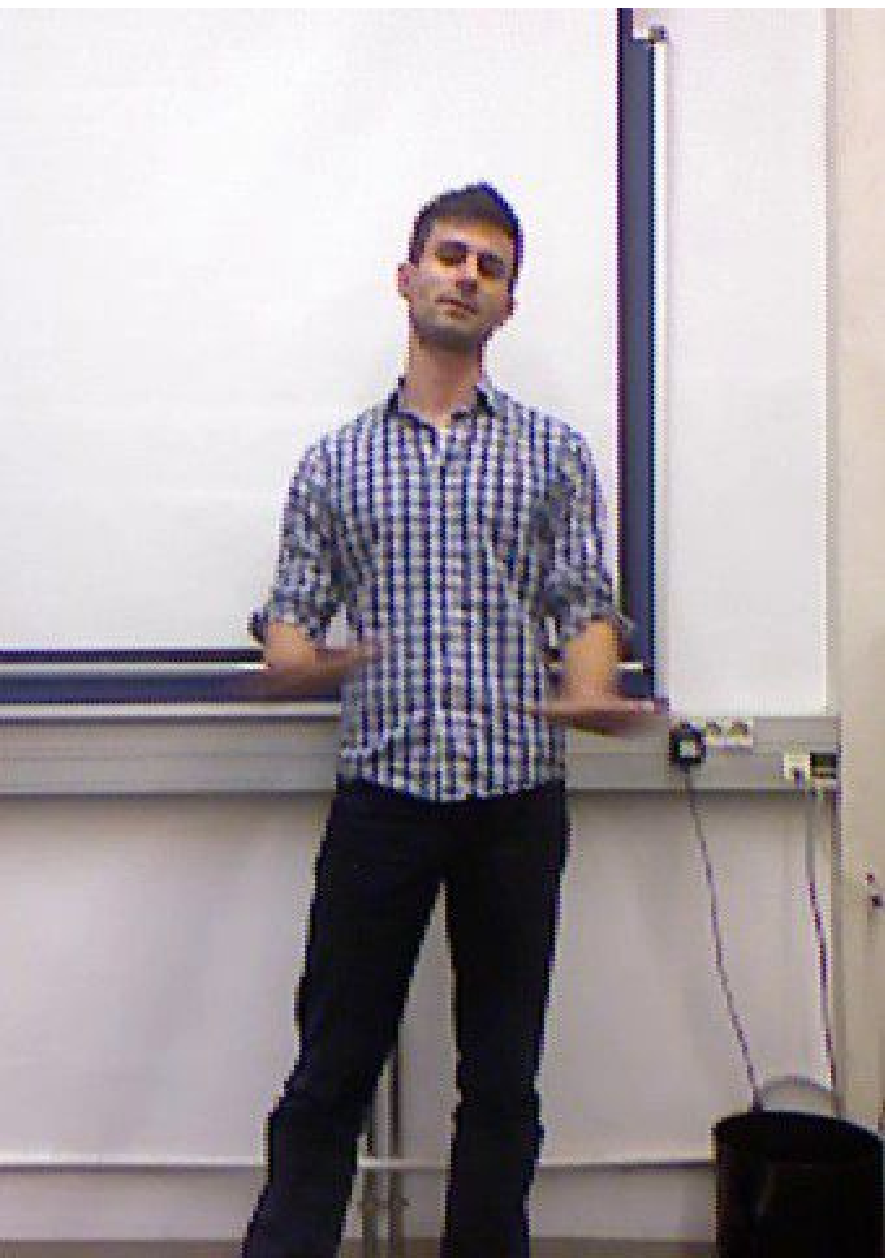}\enspace
          \includegraphics[width=0.18\textwidth]{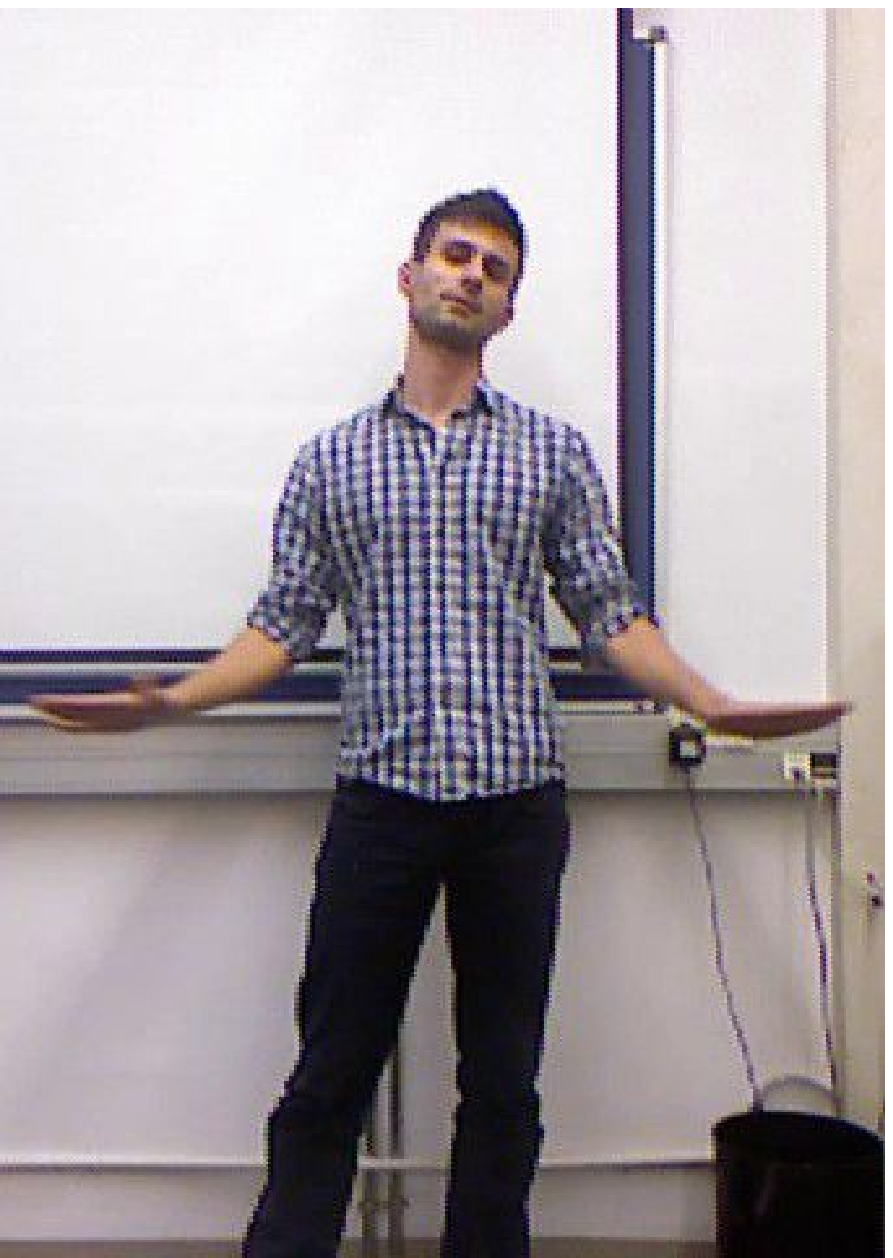}\enspace
          \includegraphics[width=0.18\textwidth]{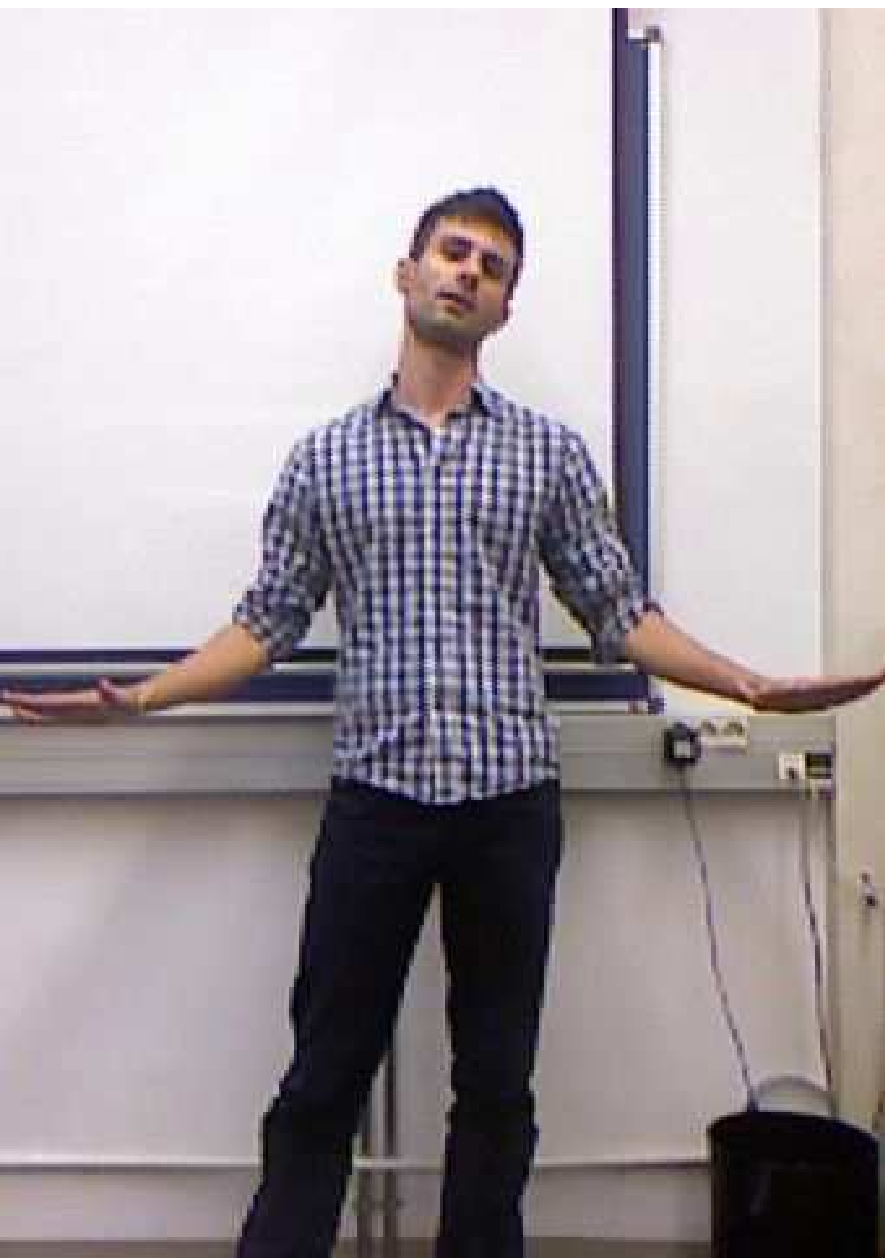}\enspace
          \includegraphics[width=0.18\textwidth]{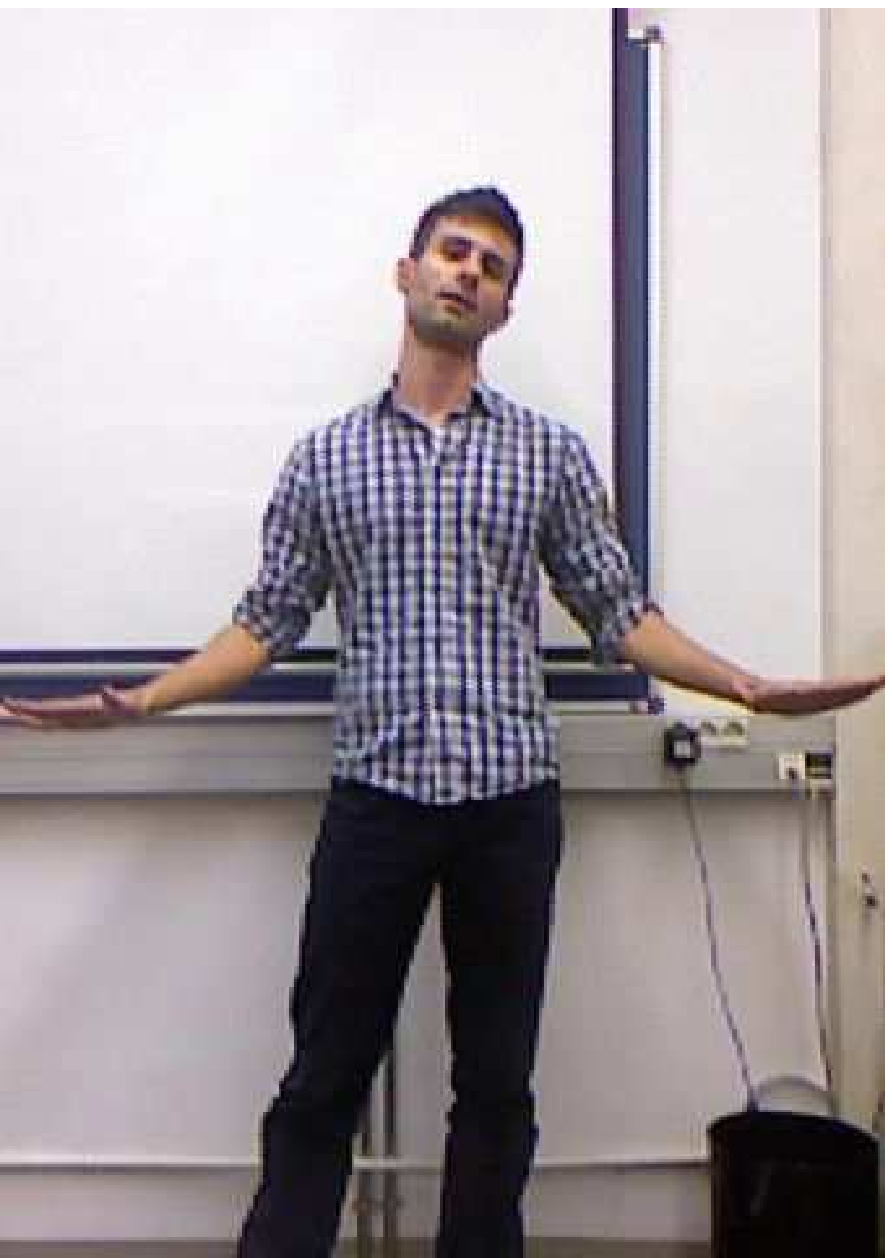}\enspace
      \end{minipage}%
    }\par
    
    \subfloat[]{%
      \begin{minipage}{\linewidth}
          \includegraphics[width=0.18\textwidth]{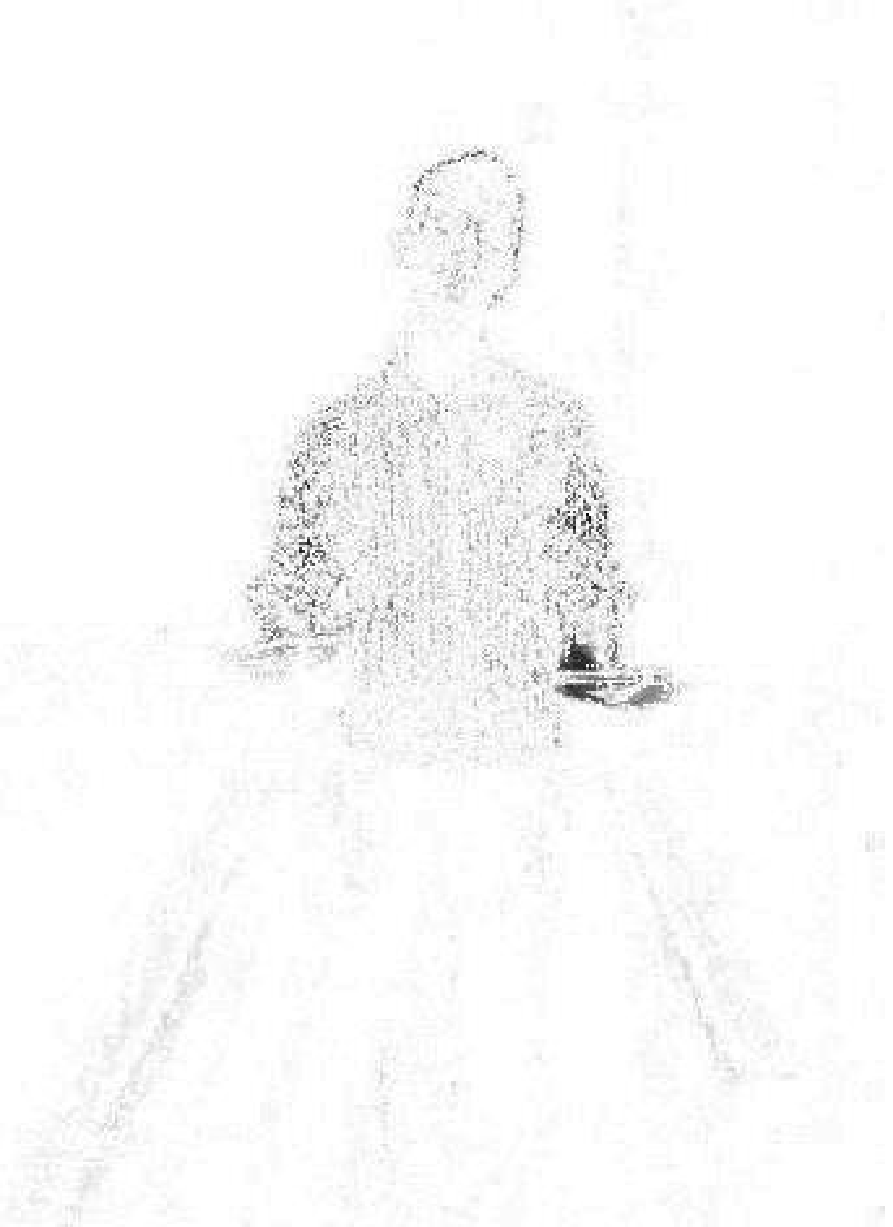}\enspace
          \includegraphics[width=0.18\textwidth]{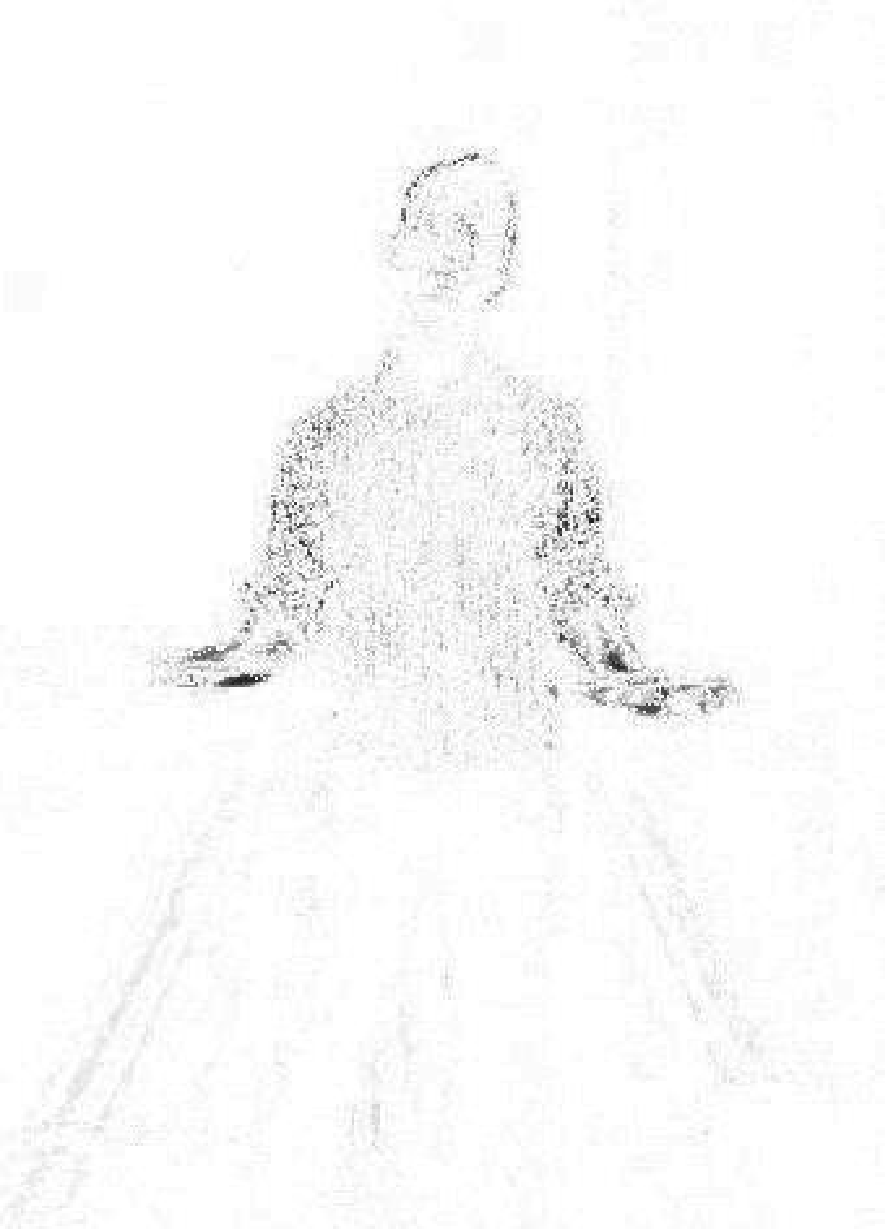}\enspace
          \includegraphics[width=0.18\textwidth]{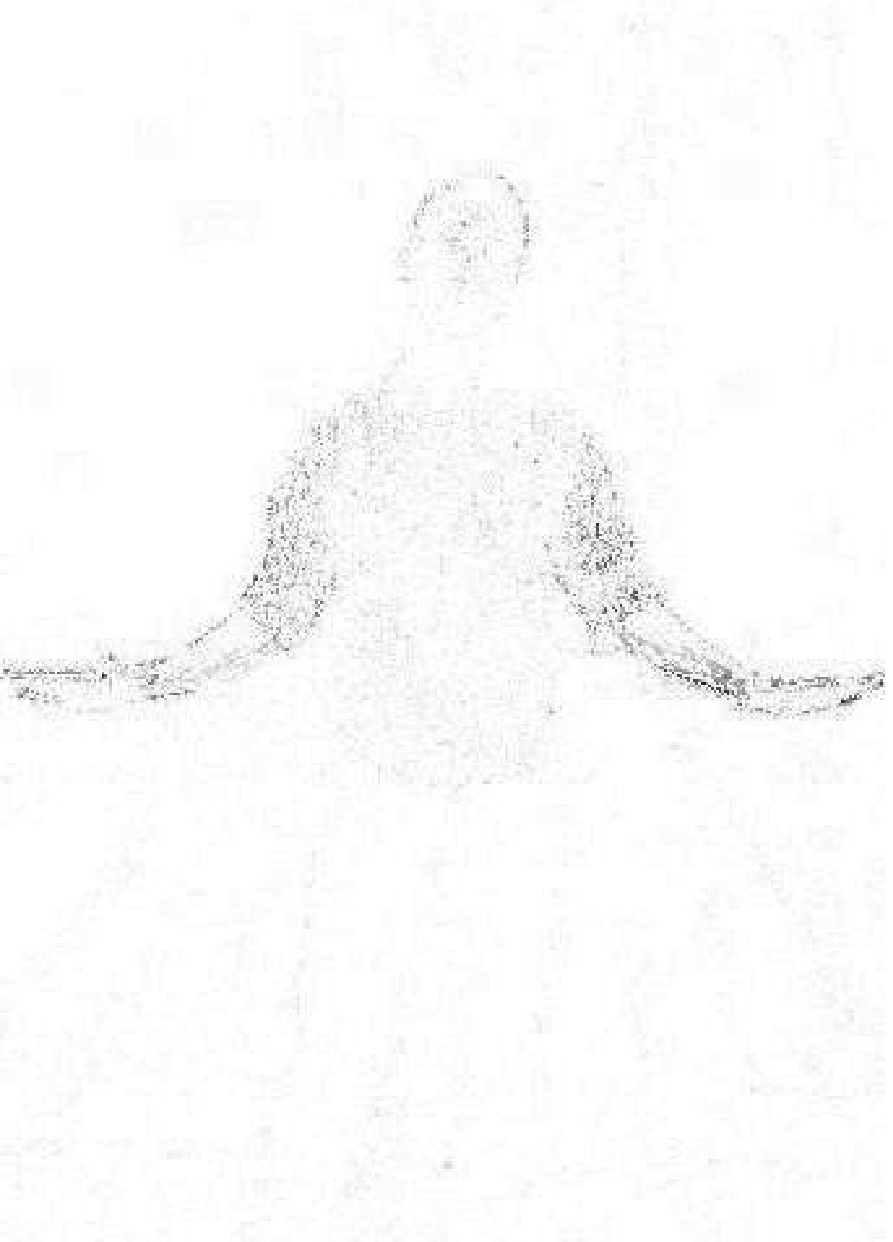}\enspace
          \includegraphics[width=0.18\textwidth]{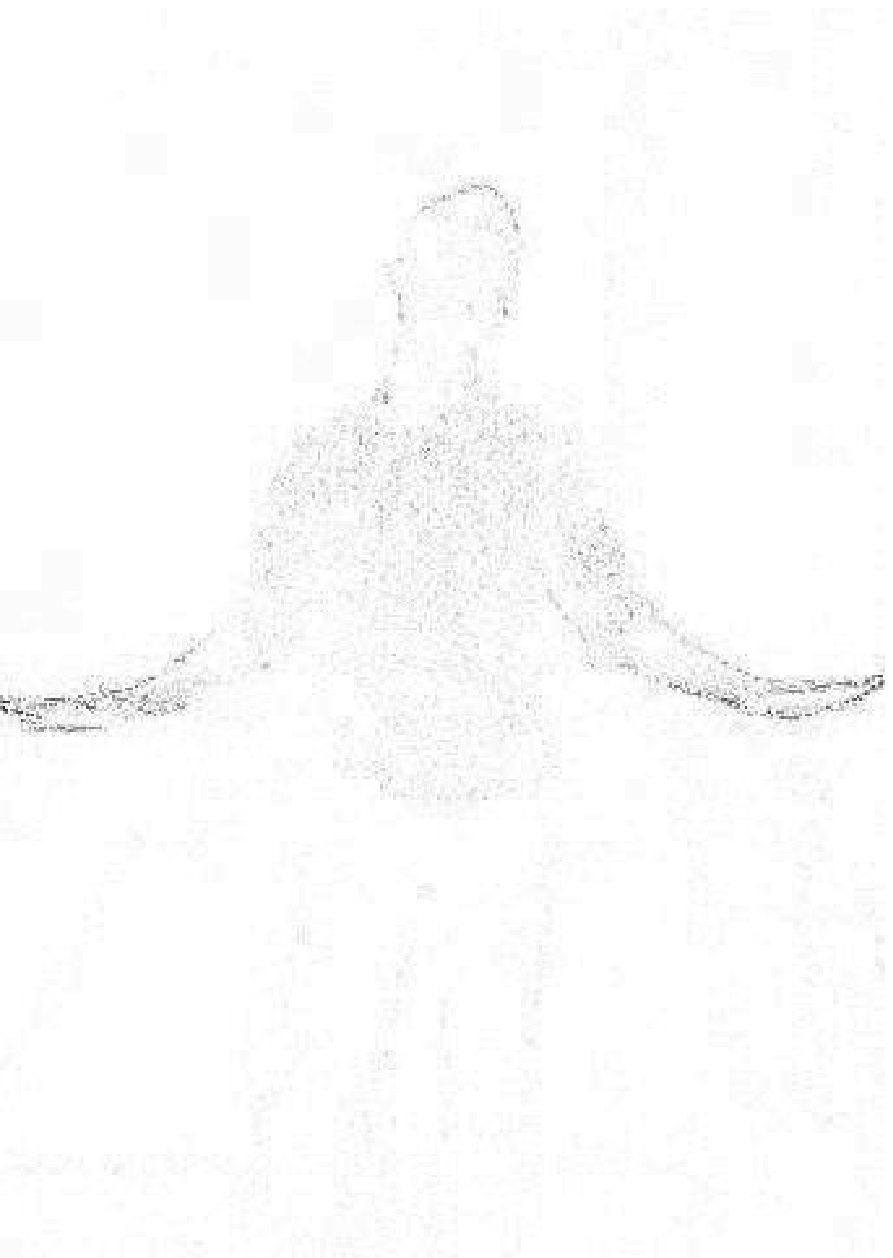}\enspace
          \includegraphics[width=0.18\textwidth]{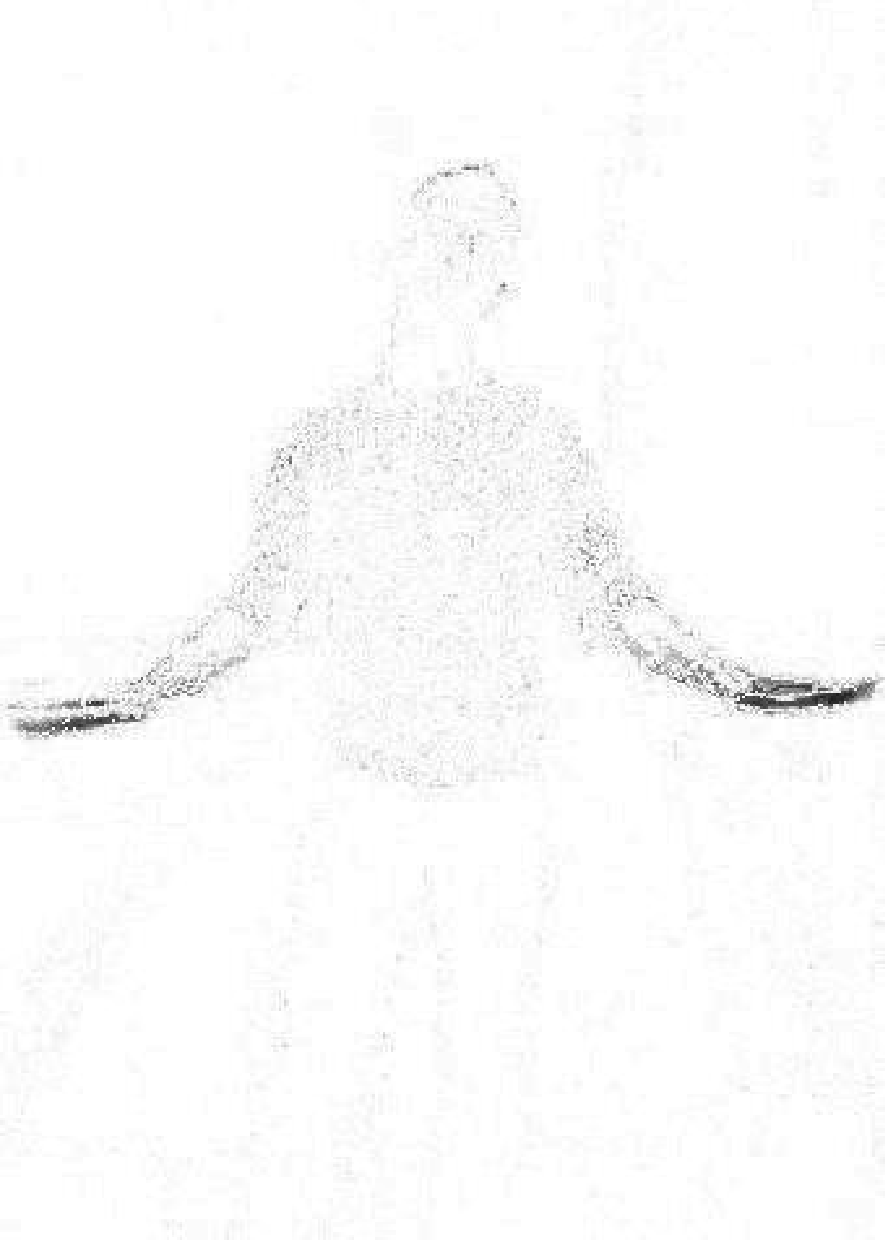}\enspace
      \end{minipage}%
    }\par
  
    \caption{A comparison between \emph{implicit} and \emph{explicit} hand movements. We observe missing hand details concerning ``sonostufo'' (b). In contrast, the \emph{explicit} arm movement of ``basta'' is conserved (d). (a) and (c) depict the original sequence for clarity. We increase the contrast of (b) and (d) for clarity.}%
    \label{fig:example_3}%
\end{figure}

\paragraph{Explicit Movements:}
Five Montalbano gesture classes require a synchronized movement of both arms. We observe two types of movements under this category according to the way the arms are extended. ``Chevuoi'' 
and ``combinato'', are performed using symmetric hand movements in which both arms move from the rest position to making a distinct shape at chest level. Due to the similarity of motion, the CNNLSTM comes short in terms of F1-scores, most noticeable for ``chevuoi''. Furthermore, \emph{Snapture} and \emph{Snapture\textsubscript{thold}} present a noticeable F1-score boost for these classes (cf. Fig.~\ref{fig:montalbano_bar_chart}). On the other hand, gestures ``cheduepalle'' and ``basta'' are also symmetric but made with a movement of both arms to the side of the body. Both gestures are used in a situation where a person is being decisive and implying ``enough''. Therefore, the movement of the arm is quite firm, making it unique from the rest of the gesture vocabulary. Consequently, the CNNLSTM is efficient at picking up these movements. \emph{Snapture} and \emph{Snapture\textsubscript{thold}} only slightly improve over the performance of the CNNLSTM concerning these \emph{explicit} movements since the handshape and finger arrangement play a minimal role in their recognition. In Fig.~\ref{fig:example_3}, we display a comparison between an \emph{implicit} and \emph{explicit} movement and their corresponding pre-processing step.

\begin{table}[]
\centering
\caption{The search space and optimal hyperparameter values (in bold) of each model in the \emph{Montalbano} experiment.}
\label{table:montalbano_hyperparameters}
\begin{tabular}{lll}
\hline
\textbf{Hyperparameter}   & \textbf{CNNLSTM}          & \textbf{Snapture$^{*}$}         \\ \hline
\textbf{Learning rate}    & {[}0.01, \textbf{0.001}, 0.0001{]} & {[}0.01, \textbf{0.001}, 0.0001{]} \\
\textbf{Number of epochs} & {[}20, 40, 60, 80, \textbf{100} {]}         & {[}20, 40, 60, 80, \textbf{100} {]}         \\
\textbf{Mini-batch size}  & {[}16, 32, \textbf{64}, 128{]}     & {[}16, 32, 64, \textbf{128}{]}     \\
\textbf{Optimizer}        & {[}\textbf{Adam}, SGD{]}           & {[}\textbf{Adam}, SGD{]}           \\ \hline
\multicolumn{3}{l}{$^{*}$\footnotesize{\textbf{Similar for \emph{Snapture\textsubscript{thold}}.}}} \\
\end{tabular}
\end{table}

\begin{table}[]
\centering
\caption{The results of the \emph{Montalbano} experiment under the described settings. The reported metrics represent the mean of five trials, while the values in parentheses correspond to the standard deviation. The superior accuracy and F1-score values are in bold.}
\label{table:montalbano_results}
\begin{tabular}{llll}
\hline
\textbf{Model}    & \textbf{CNNLSTM} & \textbf{Snapture} & \textbf{Snapture\textsubscript{thold}} \\ \hline
\textbf{Accuracy} & 0.699 (0.014)     & 0.755 (0.021)     & \textbf{0.77} (0.008)           \\
\textbf{F1-score} & 0.701 (0.013)    &  0.752 (0.021)     &  \textbf{0.772} (0.007)           \\
\textbf{Time$^{*}$}     & 234.762 (0.115)  & 318.578 (0.428)   &  744.953 (0.724)         \\ \hline
\multicolumn{3}{l}{$^{*}$In minutes.} \\
\end{tabular}
\end{table}

\section{Discussion and Future Work}
\label{sec:discussion}
In this study, we proposed a hybrid (static/dynamic) gesture recognition architecture called \emph{Snapture}. Our model integrates the hand pose alongside movement through modular static and dynamic channels. Our work is motivated by the limitation of RGB techniques, such as the CNNLSTM network, across different gesture domains. Therefore, we evaluated our approach in the context of robot commands and co-speech gestures. In our experiments, we compared the performance of both our \emph{Snapture} model and the CNNLSTM approach using the GRIT~\cite{tsironi16} and Montalbano~\cite{escalera2014} datasets. Moreover, we showed the superiority of our approach in the scope of \emph{indistinctive} and \emph{subtle} movements. Our evaluation and analysis demonstrated that considering the handshape and finger arrangement at the gesture's peak led to superior per-class F1-score values. Furthermore, we identified a lack of literature concerning the analysis of gesture motion profiles. Thus, we proposed an SSIM-based algorithm for analyzing motion profiles. We utilized this technique to propose a threshold-based extension \emph{Snapture\textsubscript{thold}}. The performance is further improved by regulating the static channel and bypassing the \emph{blurriness} issue. We believe the unique characteristics of our approach make it potentially beneficial in the following domains: 1) emblematic hand gestures, which substitute words to convey a particular meaning, and 2) co-speech gestures, which accompany words as means of verbal communication. Although we do not consider speech in our approach, incorporating additional modalities through extra channels is straightforward due to the modularity of our architecture.


Vision-based approaches are highly influenced by similarities in hand movement patterns. Furthermore, they fall short in capturing delicate small-scale hand motions at the peak~\cite{canuto20}. The effects of this phenomenon are limited in the GRIT dataset to the classes ``hello'', ``no'', and ``stop''. However, robot commands are motion-oriented, designed to be unique and convey simple meanings, i.e., robot control. On the other hand, the Montalbano Italian gestures are part of human communication. Therefore, they are natural and tend to have a basic motion path, and rather involve particular hand and finger configurations. Our analysis of the classification behavior of the CNNLSTM reports 
F1-score values of lower than 0.7 across ten classes of the Montalbano classes. This issue is also prominent in recent state-of-the-art approaches. Many of the modern gesture recognition techniques, such as 3DCNN~\cite{wu2016}, ResNet~\cite{canuto20} and Inception V3~\cite{mazhar2021} require advanced transform learning techniques and relatively long training times. In contrast, our system addresses this problem by using a simple additional static channel. Consequently, our approach facilitates a robot application due to the lightweight architecture. Furthermore, recent approaches that deal with the Montalbano dataset predominately utilize multimodalities, such as skeleton and depth data~\cite{wu2016}\cite{mazhar2021} alongside RGB, for the detection and extraction of the hand. Therefore, such approaches require a window of frames and suffer the occasional loss of joint information.
We avoid such dependencies by using RGB data only, making our approach one of the few pure RGB-based models that operate on the Montalbano dataset. Thus, it is compatible with any system equipped with a camera, including robots.

Additionally, we show that \emph{Snapture} enhances the performance by limiting the confusion between the classes that share the same movement path. Thus, it has a powerful false-positive limiting characteristic, making it a viable asset in critical scenario applications. 
One example of safety is shown in the literature in which the human operator has control over a robotic arm~\cite{mazhar2021}. The study addresses a physical safety scenario where the communication is carried out through gestures. However, their network is trained on independent static and dynamic gestures. Thus, their framework can only recognize either a static or dynamic gesture at a given time and does not address the potential risk resulting from gesture confusion. 
Therefore, this contradicts their claim of a static and dynamic gesture recognition framework. In contrast, our system integrates both the spatial and temporal aspects of the gesture and considers both for classification. Furthermore, the scheme of fusing the static and dynamic features influences the system. Our approach operates on a single frame in the static channel, which has several advantages. First, it matches with Kendon's~\cite{kendon2011} model of gesticulation and concurrent speech. It was proven in the literature that the \emph{stroke} phase plays an essential role in recognition. 
Second, the spatial and temporal traits are treated with equal importance. Thus, we can avoid the issues of fusing features at each time step. A dominance of particular modalities (RGB, depth, and skeleton) in the learning is reported by Wu et al.~\cite{wu2016}, making it more challenging to analyze the influence of each one on the final outcome. In contrast, our experimental design provides concrete evidence that classification performance concerning \emph{indistinctive} and \emph{subtle} movements can be boosted through the learning of hand details.

Furthermore, our observations on the GRIT and Montalbano datasets show high variability in hand preference. Besides hand dominance, fatigue and injuries are among the most common factors that drive the interchangeable use of both hands. Therefore, a robust system that works with subjects regardless of the dominant hand is desired. Our system accomplishes that by extracting the pose of the hand actively used when making the gesture.
That makes our system unique to other studies that mirror all videos of left-handed subjects~\cite{wu2016} or have a dedicated network for each hand~\cite{mazhar2021}. Thus, our approach facilitates higher flexibility, which leads to less restrictive and guided HRI scenarios. However, this is one step towards a wider research domain.

One of the main future directions of our work is extending the model with the body pose information. By extracting the handshape information at the peak, our architecture is prone to confusion between the classes that share a similar hand pose at the \emph{stroke} phase. Such faulty behavior can be avoided by integrating the body pose information through an additional static channel with a different cropping size. Our modular architectural design facilitates that by incorporating additional channels. 
Furthermore, gestures such as ``furbo'' and ``buonissimo'' are almost identical at the peak with minor distinction. Precise recognition of these classes depends on the context and requires additional speech or facial information. It remains interesting to extend our model with additional facial or speech features through added networks such as CNNs. Finally, one of the main disadvantages of threshold-based approaches is the lack of guarantee of generalizing to new samples. Therefore, introducing robustness by learning the cut-off values of the threshold is desirable. However, our results show that the \emph{blur} phenomena is non-trivial. We hope that our work raises more attention to the quality of collected RGB gesture datasets and encourages more research in the area of producing affordable higher-quality cameras that are compatible with robots. 



\section{Conclusion}
\label{sec:conclusion}

Despite the advantages of RGB-based vision-based hand gesture recognition frameworks, they are still challenged by the confusion between gestures with similar paths as well as the loss of hand details. In this study, we presented a novel architecture called \emph{Snapture} which integrates both the static and dynamic information of a gesture. Our RGB-only dependency and lightweight architecture design allow compatibility with any system equipped with a camera, including robots. We also suggested an algorithm for analyzing gesture motion profiles, which is essential for revealing the unique characteristics of a gesture domain. Our results provide evidence that incorporating the hand pose at the gesture's peak with motion information offers a better solution to the issues of \emph{indistinctive} and \emph{subtle} movements. They also demonstrate that these challenges are more prominent in the context of co-speech gestures compared to robot commands. Therefore, it hints at the substance of evaluating gesture recognition frameworks across multiple gesture domains. Additionally, our \emph{Snapture\textsubscript{thold}} extension highlights the influence of RGB data quality on the performance of the model and provides means for optimization based on data quality. Overall, we hope our work provides a solid step at bridging the gap between static and dynamic gestures and leading to applications that foster immersive and less controlled HRI experiences.

\begin{acknowledgements}
This work was partially supported by the DFG under project CML (TRR 169) and BMWK under project KI-SIGS. The authors would like to thank Philipp Allgeuer for the insightful comments that helped improve the first draft of this manuscript.
\end{acknowledgements}

\section*{Compliance with Ethical Standards}

The authors declare that they have no conflict of interest.

\bibliographystyle{spmpsci}      
\bibliography{ref.bib}   

%
%

\end{document}